%% file: root.tex
\documentclass[letterpaper, 10 pt, conference]{ieeeconf}  

\IEEEoverridecommandlockouts                              

\usepackage{times}
\usepackage{epsfig}
\usepackage{graphicx}
\usepackage{amsmath}
\usepackage{amssymb}
\usepackage{subcaption}
\usepackage{url}
\usepackage{booktabs}
\usepackage{multirow}
\usepackage{colortbl}
\usepackage[dvipsnames]{xcolor}

\usepackage[breaklinks=true,bookmarks=false,colorlinks,urlcolor=black, backref=page]{hyperref}
\newcommand\email[1]{\href{mailto:#1}{\nolinkurl{#1}}}

\usepackage[normalem]{ulem}
\definecolor{mygray}{gray}{0.8}
\newcommand{\clinegrayone}{\arrayrulecolor{mygray}\cline{1-10}\arrayrulecolor{black}}

\def\eg{\emph{e.g.}}

\def\etal{\emph{et al.}}

\title{\LARGE \bf
DiPE: Deeper into Photometric Errors for Unsupervised Learning of Depth and Ego-motion from Monocular Videos}

\author{Hualie Jiang, Laiyan Ding, Zhenglong Sun, and Rui Huang$^{*}$ 
\thanks{All authors are with School of Science and Engineering, The Chinese University of Hong Kong, Shenzhen, Guangdong, 518172, P.R. China. Zhenglong Sun and Rui Huang are also with Shenzhen Institute of Artificial Intelligence and Robotics for Society, Shenzhen, Guangdong, 518172, P.R. China. $^{*}$Corresponding author: Rui Huang (email: \href{mailto:ruihuang@cuhk.edu.cn}{ruihuang@cuhk.edu.cn}).} \thanks{This work is supported in part by National Key R\&D Program of China (Grant No. 2017YFB1303701), Shenzhen Natural Science Foundation (Grant No. JCYJ20190813170601651), and funding from Shenzhen Institute of Artificial Intelligence and Robotics for Society.
}
}

\begin{document}

\maketitle
\thispagestyle{empty}
\pagestyle{empty}

\begin{abstract}

Unsupervised learning of depth and ego-motion from unlabelled monocular videos has recently drawn great
attention, which avoids the use of expensive ground truth in the supervised one.
It achieves this by using the photometric errors between the target view and the synthesized views from its adjacent source views as the loss. Despite significant progress, the learning still suffers from occlusion and scene dynamics. 
This paper shows that carefully manipulating photometric errors can tackle these difficulties better.
The primary improvement is achieved by a statistical technique that can mask out the invisible or nonstationary pixels in the photometric error map and thus prevents misleading the networks. 
With this outlier masking approach, the depth of objects moving in the opposite direction to the camera can be estimated more accurately.
To the best of our knowledge, such scenarios have not been seriously considered in the previous works, even though they pose a higher risk in applications like autonomous driving.
We also propose an efficient weighted multi-scale scheme to reduce the artifacts in the predicted depth maps. Extensive experiments on the KITTI dataset show the effectiveness of the proposed approaches.
The overall system achieves state-of-theart performance on both depth and ego-motion estimation.

\end{abstract}


\section{INTRODUCTION}
The depth and ego-motion estimation is the core problem in Simultaneous Localization And Mapping (SLAM). 
Recently, Monocular Depth Estimation (MDE) attracts much attention, as it can be flexibly used in many applications, such as autonomous mobile robotics and AR/VR. 
Tracking the 6-DoF motion for a moving camera is also critical for these applications. 
Traditional supervised methods require expensively-collected ground truth, resulting in limited ability in generalization. 
By contrast, unsupervised learning from monocular videos \cite{zhou2017unsupervised} is a much more generalizable solution.

\begin{figure}[t]
\vspace{5pt}
\begin{center}
\includegraphics[width=0.76\linewidth]{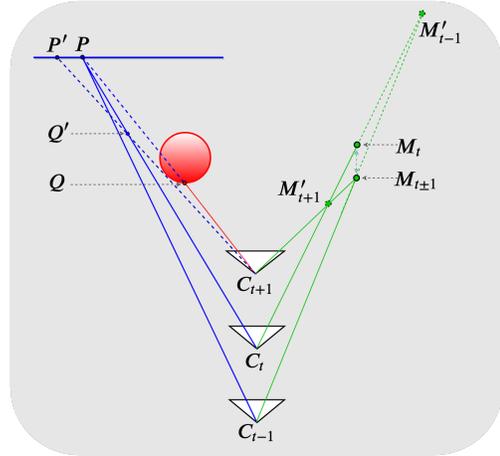}
\end{center}
\vspace{-5pt}
\caption{\textbf{The detriment of occlusion and dynamics.} \textbf{(a) Occlusion:} point $P$ is visible in $C_t$ (camera view in time $t$) but occluded in $C_{t+1}$, and to achieve photometric consistency (matching $P'$ instead of $Q$), $P$ is estimated with shorter depth ($Q'$), making the foreground object blur. \textbf{(b) Co-directional motion:} if point $M$ moves forward like camera $C$ (from $t-1$ to $t$), it is likely estimated with farther depth ($M'_{t-1}$), producing `dark holes'. \textbf{ (c) Contra-directional motion:} when point $M$ moves backward opposite to camera $C$ (from $t$ to $t+1$), it is estimated with shorter depth ($M'_{t+1}$).}
\label{fig:om}
\vspace{-10pt}
\end{figure}

The unsupervised learning models usually contain two networks for predicting the depth map of the target view, and the motion between the target view and its temporally adjacent views. 
With the network output, the target view can be reconstructed by the adjacent source views with image warping, and the resulted photometric loss can be used as the supervisory signal for learning. 
However, the image reconstruction is usually destroyed by between-view occlusion and scene dynamics, as illustrated in Fig.~\ref{fig:om}, and the resulting incorrect supervision harms the network learning.

The theory of how minimizing between-view reconstruction errors affects the depth estimation of occluded regions and the common forward and backward moving objects is illustrated in Fig.~\ref{fig:om}. 
Many methods have been proposed to cope with the occlusion and dynamics, and considerable improvement has been made. 
For example, the effect of `dark holes' by the co-directionally moving objects has been tackled in the latest work \cite{luo2019every, casser2019struct2depth, godard2019digging}. However, as shown in Fig.~\ref{fig:comparison} the latest models make significant underestimation of the depth for the contra-directionally moving objects. To the best of our knowledge, the inaccuracy of such objects has not been reported in the literature, which may cause trouble in practical applications. For instance, in autonomous driving, if the distance of oncoming cars is underrated, unnecessary braking or avoiding may be executed.

\begin{figure*}[t]
\begin{center}
\vspace{6pt}
\includegraphics[width=0.9\linewidth]{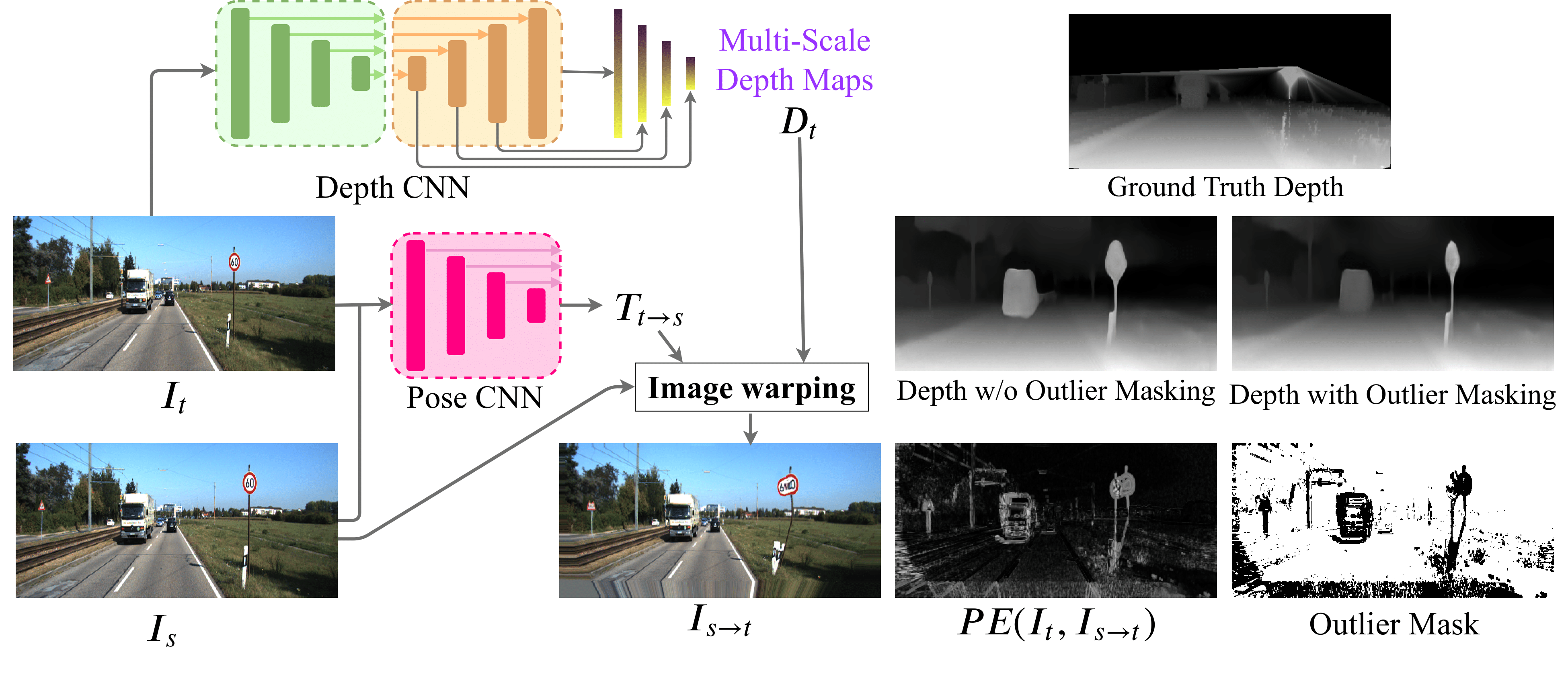}
\end{center}
\vspace{-12pt}
\caption{\textbf{The Unsupervised Learning Flow and Effect of Outlier Masking.} \textbf{(a)} {Depth CNN:} A standard fully convolutional U-net that predicts the multi-scale depth maps for the target image. {Pose CNN:} A standard CNN that inputs the target view and one source view and predicts their relative motion. With $D_t$ and $T_{t \to s}$ by the networks, the synthesized image $I_{s \to t}$ from the source view $I_s$ to the target view $I_t$ be differentiablly warped. The photometric errors between $I_t$ and $I_{s \to t}$ can work as the training objective for both the Depth CNN and Pose CNN. 
\textbf{(b)} The outlier masking can exclude many invisible and nonstatic pixels, particularly those belonging to contra-moving objects, thus predicting a more accurate depth map. Without outlier masking, the oncoming vehicle is predicted to be very close, and the foreground object boundary significantly dilates.}
\label{fig:framework}
\vspace{-10pt}
\end{figure*}

This issue can be largely avoided by our proposed {outlier masking} technique, which helps to exclude the occluded and moving regions, especially the oncoming objects. 
The technique is driven by our observation that the photometric errors of occluded and dynamic regions are much larger. 
In theory, the visible background usually dominates the scenes and the invisible or moving pixels are inconsistent with the background, thus making their errors difficult to optimize. 
Besides, we also propose an efficient weighted multi-scale scheme to reduce artifacts and work with the outlier masking to produce better depth maps. 

The effectiveness of our two main contributions, as mentioned above, is experimentally proven on the driving KITTI dataset.  Together with a simple baseline model and some other masking practices, we build an overall state-of-the-art unsupervised monocular depth and ego-motion estimation system, called DiPE.

\section{RELATED WORK}
In recent years, deep Convolutional Neural Networks (CNN) have boosted the performance of MDE. 
One typical approach is using a deep CNN to densely regress the ground truth depth obtained with physical sensors \cite{eigen2014depth, laina2016deeper, zeng2017geocuedepth, jiang2019high}. 
Other approaches can be categorized as combining deep learning with graphical models \cite{wang2015towards, liu2015learning, xu2017multi} or casting MDE as a dense classification problem \cite{cao2017estimating, li2018monocular, fu2018deep}. 
However, models trained on publicly available datasets with ground truth depth, like the NYUDepthV2 \cite{silberman2012indoor} or KITTI \cite{geiger2013vision}, usually do not generalize well to real scenarios.

Instead of depending on ground truth, unsupervised learning schemes adopt more available resources, the stereo images \cite{garg2016unsupervised, godard2017unsupervised} or adjacent monocular video frames \cite{zhou2017unsupervised} to construct the supervisory signal. Specifically, the loss is the photometric difference between a view and its synthesis. The synthesis can be computed from the additional view by its estimated depth and the known or estimated pose between the two views. To take advantage of both spatial and temporal cues, stereo videos are exploited for training in \cite{zhan2018unsupervised, li2018undeepvo, luo2019every, godard2019digging}. Compared with stereo images, the monocular videos are more generalized and available, thus this paper focuses on the latter one.  

The first method training with monocular videos, SfMLearner \cite{zhou2017unsupervised} adopts an additional Pose CNN to estimate the relative motion between sequential views to make view synthesis attainable. However, the photometric consistency between nearby views is usually unsatisfied due to occlusion and moving objects. 
To improve this advantageous framework, many methods have been proposed, which can be mainly classified as: masking photometric errors \cite{zhou2017unsupervised, godard2019digging, klodt2018supervising, wang2019unsupervised}, joint learning with optical flow \cite{luo2019every, yin2018geonet, zou2018df}, modelling object motion \cite{casser2019struct2depth, vijayanarasimhan2017sfm, Gordon_2019_ICCV}. 
The masking strategies also do not necessarily guarantee flexibility. 
Some masking techniques, such as the explainability mask \cite{zhou2017unsupervised} and the uncertainty map \cite{klodt2018supervising} also requires an extra network to learn. 
Joint learning with the optical flow has to construct a new network for learning optical flow to explain or compensate for the photometric inconsistency caused by occlusion or scene dynamics. 
Similarly, modelling object motion also requires additional modules to estimate the segmentation and motion of objects.

Different from the above methods, the overlap and blank masks geometrically derived from the image warping process \cite{wang2019unsupervised} is a light-weight design for occlusion. A simpler method for occlusion is the {minimum reprojection} in Monodepth2 \cite{godard2019digging}, which takes the minimum photometric errors from all source views, thus is also a masking technique. Monodepth2 also adopts a {auto-masking} technique for moving objects in a close speed with the camera. This simple and efficient masking strategy has been proved effective by Monodepth2, compared with other state-of-the-art methods. However, the oncoming moving objects, have not been noticed and solved. The outlier masking method is proposed in this paper for such objects. Further, our {outlier masking} technique can help the {minimum reprojection} to recover a more accurate boundary for the foreground objects in predicted depth maps. 


\section{METHODOLOGY}
\subsection{Preliminaries}
The monocular unsupervised learning scheme is shown in Fig.~\ref{fig:framework}.
A training sample contains the target frame $I_t$ at time $t$ and some source frames $I_s$ at nearby times, $s \in \mathcal{S}$. Conventionally,  $\mathcal{S} = \{t-1, t+1\}$ or $\{t-2, t-1, t+1, t+2\}$. Suppose that $K$ is the shared intrinsic matrix of these frames. With the predicted depth $D_t$ and transformation $T_{t, s}$, the synthesis from the source view $s$ to the target view $t$ can be expressed as,
\begin{equation}
\quad I_{s \to t} = I_{s}\langle proj(D_t, T_{t \to s}, K) \rangle, 
\label{eq-synth}
\end{equation}
where $\big\langle\big\rangle$ is the differentiable bilinear sampling operator \cite{jaderberg2015spatial} and $proj()$ is the operation projecting the pixel $p_t$ in the target image to the point $p_s$ in the source image, 
\begin{equation}
p_s \simeq KT_{t \to s}D(p_t)K^{-1}p_t,
\label{eq:proj}
\end{equation}
where $p_t$ and $p_s$ are expressed in homogeneous coordinates. 

In this paper, we adopt the popular  combination of L1 and SSIM by \cite{godard2017unsupervised} to compute the photometric errors, 
\begin{equation}
\mathcal{PE}(I_a, I_b) = 0.85 \frac{1 - \mathrm{SSIM}(I_a, I_b)}{2} + 0.15 \|I_a - I_b\|_1,
\end{equation}

In addition, an edge-aware smoothness term is usually also applied in unsupervised training.  We use the one by \cite{godard2019digging},
\begin{equation}
L_{es} = mean\left( \left | \partial_x d^*_t   \right | e^{-\left | \partial_x I_t \right |} + \left | \partial_y d^*_t   \right | e^{-\left | \partial_y I_t \right |} \right), 
\label{eq:smooth}
\end{equation} 
where $d^*_t = d_t / \overline{d_t}$ is the mean-normalized inverse depth from \cite{wang2018learning} to discourage shrinking of the estimated depth. Both losses are applied in $4$ scales to avoid gradient locality.

\subsection{Outlier Masking}
As has been discussed, the image reconstruction can be damaged by some adverse factors, such as occlusion and scene dynamics. Therefore a portion of pixels in the photometric error map is invalid, and the incorporation of them in training can be misleading. 
We have the observation that most pixels are visible and stationary, and other occluded and moving pixels always produce more significant photometric errors. 
The outlier masking technique is based on this observation, which is simple but effective. The outlier mask is automatically determined by the statistical information of photometric errors. 
Specifically,, we first compute the mean and standard deviation of pixel photometric errors from all source images for every training sample, 
\begin{eqnarray}
\mu &=& mean{ \{\mathcal{PE}(I_t, I_{s \to t}) | s \in \mathcal{S}\}}, \label{eq:pemean}\\
\sigma &=& std{ \{\mathcal{PE}(I_t, I_{s \to t}) | s \in \mathcal{S}\}}. \label{eq:std}
\end{eqnarray}
Then, we compute an outlier mask for the photometric error map $\mathcal{PE}(I_t, I_{s \to t})$, 
\begin{equation}
M_s^{ol} =  \mu - l\sigma<\mathcal{PE}(I_t, I_{s \to t}) < \mu + u\sigma, 
\label{eq:Mol}
\end{equation} 
where $l$ and $u$ are the lower and upper thresholds. 

\begin{figure}[t]
\vspace{6pt}
\begin{center}
\includegraphics[width=\linewidth]{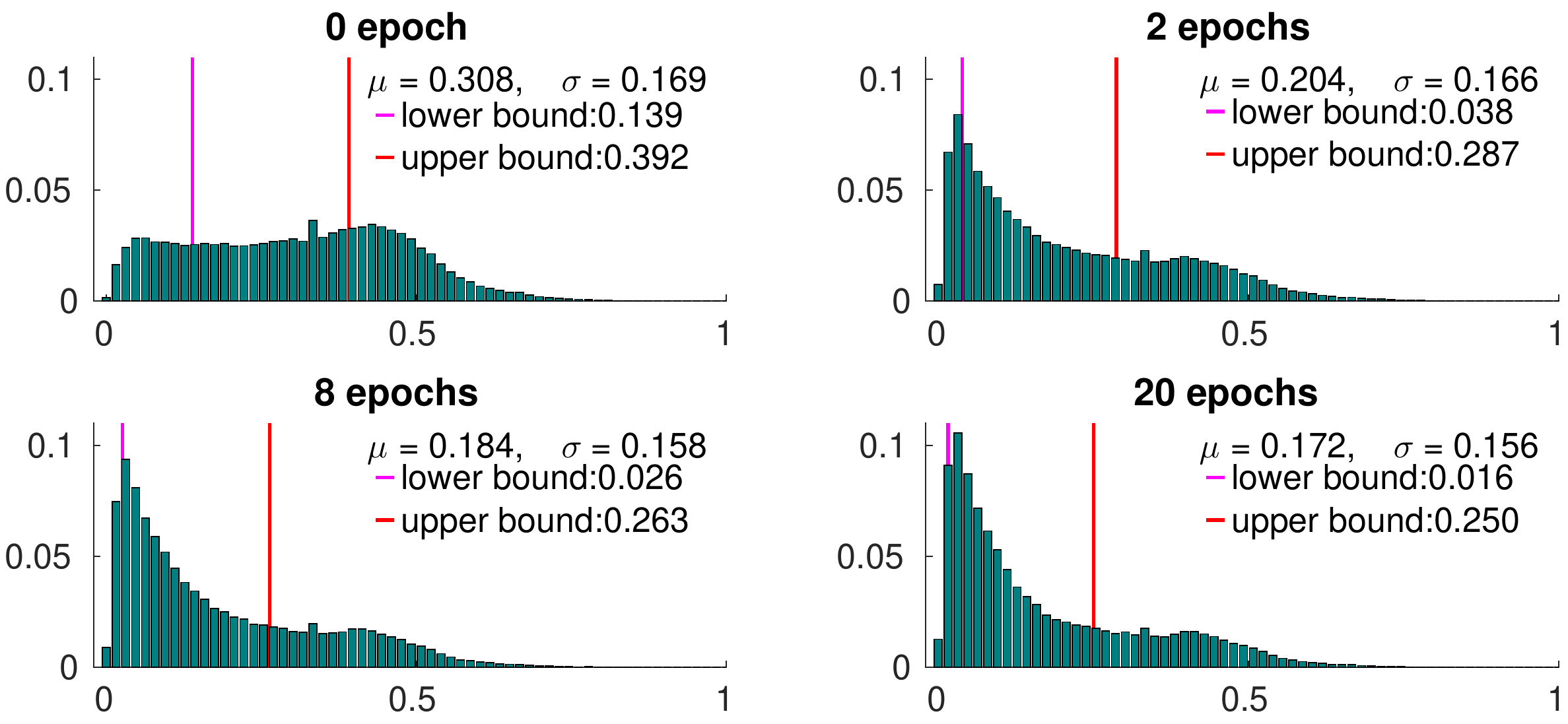}
\end{center}
\vspace{-6pt}
\caption{\textbf{The variation of the photometric error distribution.} The photometric error distribution of a validation sample changes during training.  Before training, even with a little long tail, the errors are distributed somewhat evenly. After 2 epochs, the majority of errors converge to the lower bound, and a notable long tail forms. Next, the errors under the upper bound continue to decrease and converge, but the errors in the long tail do not change much. }
\label{fig:pe}
 \vspace{-8pt}
\end{figure}

We can use the computed mask to exclude the possible occluded or moving regions, as shown in Fig.~\ref{fig:framework}. By visualizing the resulted masks during training, we find that it is good to set $u$ as 0.5 because a higher value cannot sufficiently mask the moving objects, and a lower value can mask out many stationary objects. Besides, $l$ is set as $1$ to mask some pixels with very small photometric errors because these pixels usually belong to homogeneous regions and not very valuable for network training. This selection for $u$ and $l$ can retain the principal photometric errors for optimizing, as shown in Fig. ~\ref{fig:pe}.

\subsection{Weighted Multi-Scale Scheme}
To avoid getting stuck in local minima due to the gradient locality of the bilinear sampler \cite{jaderberg2015spatial}, the unsupervised learning models usually predict 4 scale depth maps (Fig.~\ref{fig:framework}) and compute multi-scale photometric losses for training. 
However, it has been pointed out that this scheme tends to produce `holes' in large low-texture regions in the intermediate lower resolution depth maps, as well as texture-copy artifacts \cite{godard2019digging}. 
To alleviate this phenomenon, Monodepth2 \cite{godard2019digging} adopts a full resolution multi-scale scheme, i.e., to upsample the multi-scale depth maps to the full resolution, perform the image warping using the full-resolution images, and compute photometric losses at the full resolution. 

However, we find that this full-resolution scheme considerably increases the computation and GPU memory during training.  
To suppress the phenomenon without raising training overhead, we propose a weighted multi-scale scheme to devalue the low-resolution photometric losses and lighten the disadvantage they bring. 
Explicitly, we define a scale factor $f<1$ to compute the weight for the scale $r$, 
\begin{equation}
w_r =    f^r, 
\label{eq:scale-weight}
\end{equation} 
where $r \in \{0, 1, 2, 3\}$.

\subsection{Integrated Objective Function}
\label{sec:objective}
Although the proposed outlier masking technique can exclude most irregular pixels, it has some failure cases. 
For example, the outlier masking cannot eliminate the pixels that move out of the image boundary, as illustrated by the bottom of the outlier mask in Fig.~\ref{fig:framework}. 
In fact, it is easy to mask the out-of-box pixels by the principled masking technique \cite{mahjourian2018unsupervised}, which only retains the pixels that are reprojected inside the image box of the source images. 
Besides, the outlier masking cannot mask out the objects with a very close speed to the camera, as these objects are usually estimated to the maximum depth, and the corresponding photometric errors can exactly lie in the statistical inlier region. 
As illustrated in Fig.~\ref{fig:framework}, the car in the same lane is not well masked in the outlier mask. Fortunately, the auto-masking technique in Monodepth2 \cite{godard2019digging} can handle such cases and after including this masking method, the car in the front is not estimated to be very far away. Moreover, we also find that our outlier masking can collaborate with the minimum reprojection in \cite{godard2019digging} well to produce more accurate foreground object boundaries in the predicted depth maps. 
Therefore, we build a baseline with these three techniques. 

The auto-masking excludes the pixels that hold larger photometric errors by reconstruction than the direct photometric error between the target view and the source view. The minimum reprojection is also a masking technique, and it only retains the pixels with the minimum photometric error among all source views. We express the masks of these three masking methods for the photometric error map $\mathcal{PE}(I_t, I_{s \to t})$ as $M_s^{p}$, $M_s^{a}$ and $M_s^{mr}$, 
\begin{eqnarray}
M_s^{p} &=& [p_s  \text{ within image box}], \label{eq:Mp} \\
M_s^{a} &=& \mathcal{PE}(I_t, I_{s \to t}) < \mathcal{PE}(I_t, I_s), \label{eq:Ma} \\
M_s^{mr} &=& \mathcal{PE}(I_t, I_{s \to t}) \leq  \min_s{\mathcal{PE}(I_t, I_{s \to t})}, \label{eq:Mmr}
\end{eqnarray}
where $p_s$ is calculated by Eqn.~\ref{eq:proj}. 
Then we can compute the final mask for the photometric error map $\mathcal{PE}(I_t, I_{s \to t})$ by combining three type masks, 
\begin{equation}
M =  M_s^{ol} \bullet M_s^{p} \bullet M_s^{a} \bullet M_s^{mr} ,
\label{eq:Mask}
\end{equation} 
where $\bullet$ represents the element-wise logical conjunction. 
Finally, the overall objective function is computed by,
\begin{equation}
L = \eta \sum_{r} f^r \sum_{s} \frac{M_sP_s}{\#\{M_s=1\}} +\lambda\sum_{r} e^rL_{es}^r, 
\label{eq:loss}
\end{equation} 
where we denote $\mathcal{PE}(I_t, I_{s \to t})$ as $P_s$, $\eta$ and $\lambda$ are weights to balance the two types of losses, and $e$ is a weighting factor for the edge-aware smoothness loss from different scales. 

\section{EXPERIMENTS}

\begin{table*}[t]
\vspace{5pt}
  \centering
  \resizebox{0.88\textwidth}{!}{
  \begin{tabular}{l|c|cccc|ccc}
  \toprule
  \multirow{2}{*}{Method} & \multirow{2}{*}{Train} & \multicolumn{4}{c|}{Error metric $\downarrow	$} & \multicolumn{3}{c}{Accuracy metric $\uparrow	$}\\
  \cline{3-6}
  \cline{7-9}
   &   & Abs Rel & Sq Rel & RMSE  & RMSE log & $\delta < 1.25 $ & $\delta < 1.25^{2}$ & $\delta < 1.25^{3}$  \\
  \hline
  \hline
Eigen~{\etal}~\cite{eigen2014depth} & D & 0.203 & 1.548 & 6.307 & 0.282 & 0.702 & 0.890 & 0.890\\
Liu~{\etal}~\cite{liu2015learning} & D & 0.201 & 1.584 & 6.471 & 0.273 & 0.680 & 0.898 & 0.967\\
Kuznietsov~{\etal}~\cite{kuznietsov2017semi} & DS & 0.113 & 0.741 & 4.621 & 0.189 & 0.862 & {0.960} & {0.986}\\
DORN~\cite{fu2018deep} & D & \textbf{0.072}&  \textbf{0.307} & \textbf{2.727} & \textbf{0.120} & \textbf{0.932} & \textbf{0.984} & \textbf{0.994}\\ 
\hline
\hline
Garg~\cite{garg2016unsupervised} & S  &  0.152 & 1.226 & 5.849 & 0.246 & 0.784 & 0.921 & 0.967\\
Monodepth R50~\cite{godard2017unsupervised}\textdagger & S & 0.133 & 1.142 & 5.533 & 0.230 & 0.830 & 0.936 & 0.970\\
SuperDepth~\cite{pillai2019superdepth} & S & 0.112 & 0.875 & \textbf{4.958} & \textbf{0.207} & 0.852 & 0.947 & \textbf{0.977}\\
Monodepth2~\cite{godard2019digging}  &  S & 
\textbf{0.109} & \textbf{0.873} &  {4.960} &   {0.209} &   \textbf{0.864} &   \textbf{0.948} &  {0.975} \\
\hline
\hline
SfMLearner~\cite{zhou2017unsupervised}\textdagger & M & 0.183 & 1.595 & 6.709 & 0.270 & 0.734 & 0.902 & 0.959\\
Vid2Depth~\cite{mahjourian2018unsupervised} & M & 0.163 & 1.240 & 6.220 & 0.250 & 0.762 & 0.916 & 0.968\\
DF-Net~\cite{zou2018df} & M & 0.150 & 1.124 & 5.507 & 0.223 & 0.806 & 0.933 & 0.973\\
GeoNet~\cite{yin2018geonet}\textdagger & M  & 0.149 & 1.060 & 5.567 & 0.226 & 0.796 & 0.935 & 0.975\\
DDVO~\cite{wang2018learning} & M  & 0.151 & 1.257 & 5.583 & 0.228 & 0.810 & 0.936 & 0.974\\
EPC++~\cite{luo2019every} & M & 0.141 & 1.029 & 5.350 & 0.216 & 0.816 & 0.941 & 0.976\\
Struct2depth `(M)'~\cite{casser2019struct2depth}  & M & 0.141 & {1.026} & 5.291 &  0.215 & 0.816 & 0.945 & {0.979}\\
Gordon~\etal \cite{Gordon_2019_ICCV} & M & 0.128 & 0959 & 5.230 &  0.212 & 0.845 & 0.947 & {0.976}\\
Monodepth2~\cite{godard2019digging} & M & {0.115} & {0.903} & {4.863} &  {0.193} &   {0.877} & {0.959} &   {\bf  0.981} \\ 
\textbf{DiPE} (Ours) & M &   
 {\bf 0.112} &   {\bf 0.875} &   {\bf 4.795} &   {\bf 0.190} &   {\bf 0.880} &   {\bf 0.960} &   {\bf 0.981} \\ 
  \bottomrule
  \end{tabular}
  }
\caption{\textbf{Quantitative Results.} All the methods are trained and evaluated on the Eigen split \cite{eigen2014depth} of the KITTI dataset \cite{geiger2013vision}. Three categories of methods, which perform training with the depth, stereo images. and monocular video frames, respectively, are compared. In each category, the best results are in \textbf{bold}. \textbf{Legend:} D -- depth supervision; S -- unsupervised stereo supervision; M -- unsupervised mono supervision; \textdagger -- newer results from the respective online implementations. }
\label{tab:kitti_eigen}
\end{table*}

\begin{figure*}[!ht]
\centering
\resizebox{\textwidth}{!}{
\input{figs/comparison/comparison.tex} }
\caption{\textbf{Qualitative comparison.} Our model DiPE produces very high-quality depth maps, and it reduces most artifacts due to occlusion and scene dynamics. More importantly, recent models, for instance, joint learning with optical flow \cite{yin2018geonet, luo2019every}, \eg oncoming cars, including modelling object motion \cite{casser2019struct2depth} and auto-masking moving objects \cite{godard2019digging}, underestimate the depth for the objects moving in an opposite direction, \eg oncoming cars,  while our DiPE succeeds (the second column). Best viewed in color and zoomed in.}
\vspace{-8pt}
\label{fig:comparison}
\end{figure*}
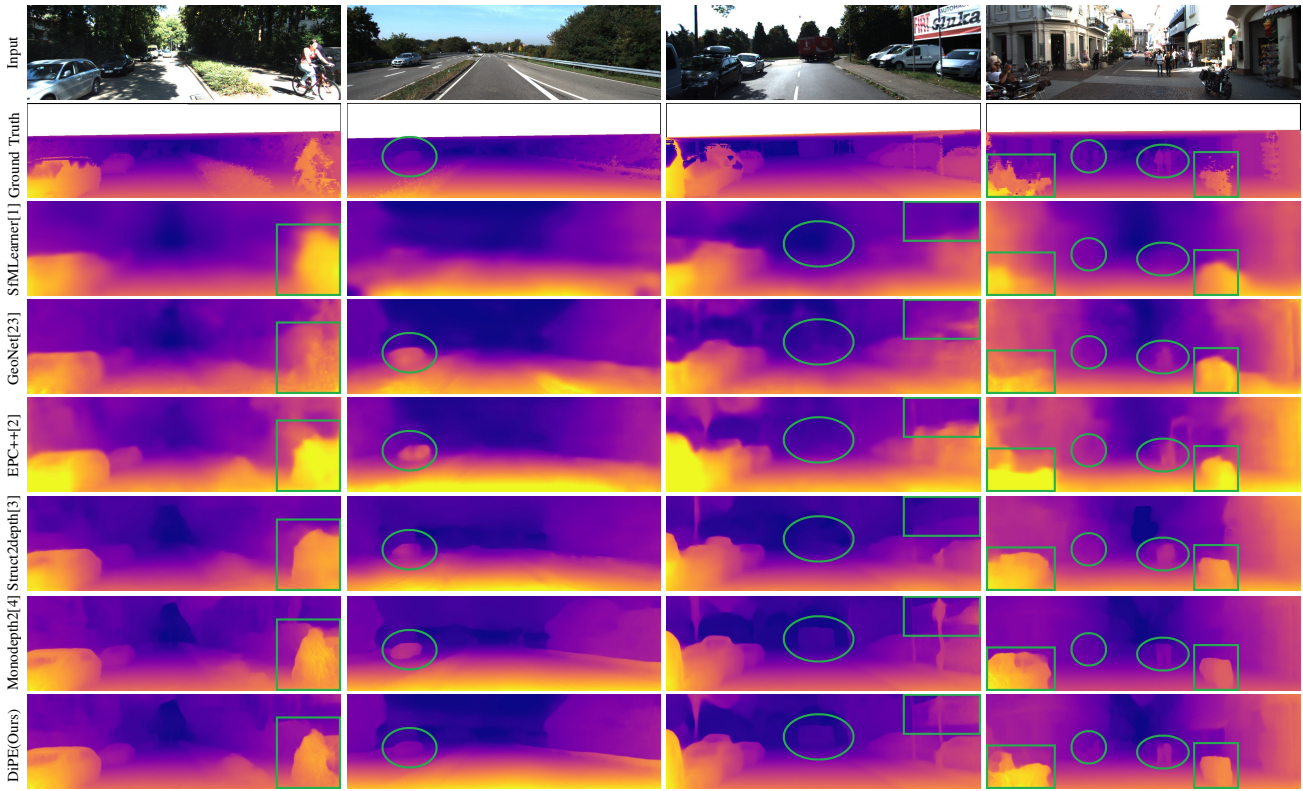

\subsection{Implementation Details}
We implement the proposed approaches based on Monodepth2 \cite{godard2019digging} and maintain the most basic experimental settings. 
The depth CNN is a fully convolutional encoder-decoder network with an input/output resolution of $640\times192$. The Pose CNN is a stand CNN with fully connected layer to regress the 6-Dof relative camera motion. 
Both networks use a ResNet18 \cite{he2016deep} pretrained on ImageNet \cite{deng2009imagenet} as backbone for all of the experiments.
In depth estimation experiments, as Monodetph2 \cite{godard2019digging}, we only use the nearby $2$ frames ($\mathcal{S} = \{t-1, t+1\}$) and the pair-input Pose (Fig.~\ref{fig:framework}). In ego-motion estimation, however, we also experiment with the all-input Pose CNN with the nearby $4$ frames  ($\mathcal{S} = \{t-2, t-1, t+1, t+2\}$) and $2$ frames ($\mathcal{S} = \{t-1, t+1\}$) for \cite{zhou2017unsupervised, mahjourian2018unsupervised}.

The hyper-parameter $\eta$, $\lambda$, and $e$ in the final loss function are empirically set to $1$, $0.001$, and $0.5$. The factor $f$ of the weighted multi-scale scheme is chosen as $0.25$ by examining several values in the validation set. 
DiPE is also trained for 20 epochs using Adam \cite{kingma2014adam}. As our weighted multi-scale scheme consumes less memory, DiPE is trained with a bigger batch size of 16 than 12 in Monodepth2 and the training spends only 9 hours on a single Titan Xp while Monodepth2 uses 12 hours. DiPE also uses an initial learning rate of  $10^{-4}$ but divides it by 5 after 15 and 18 epochs, whereas Monodepth2 divides it by 10 only after 15 epochs. As the outlier masking further reduces the errors for training and decreasing the learning rate can help DiPE converges better. 
Monodepth2 uses the same intrinsic parameters for all training samples by approximating the principal point of the camera to the image center and averaging the focal length on the whole dataset. More precisely, we use the calibrated intrinsic parameters for every training samples, and when performing horizontal flips in data augmentation, the horizontal coordinate of the principal point is also flipped. 

\subsection{KITTI Eigen Split}
We adopt the standard Eigen split \cite{eigen2014depth} of the KITTI dataset \cite{geiger2013vision} in the monocular depth estimation experiments. 
Following Zhou~\etal~\cite{zhou2017unsupervised}, we use a subset of the training set that contains no static frames for training. There are 39,810, 4,424, and 697 samples for training, validation, and test. 
We also only use about one-tenth (432) of the validation set for validation, because we evaluate all the validation samples after every epoch rather than evaluate a batch of validation samples for certain steps, and this is better for monitoring the training process without spending too much time on validation. 
In evaluation, every predicted depth map is aligned to the ground truth depth map by multiplying the median value ratio \cite{zhou2017unsupervised} as other unsupervised monocular methods, and we also adopt the conventional metrics and cropping region in \cite{eigen2014depth}, and the standard depth cap $80m$ \cite{godard2017unsupervised}. There are 4 error metrics, namely, absolute relative error (Abs Rel), square relative error (Sq Rel), root mean square error (RMSE) and the root mean square error in log space (RMSE log). Other 3 accuracy metrics are the percentages of pixels where the ratio ($\delta$) between the estimated depth and ground truth depth smaller than $1.25$, $1.25^2$ and $1.25^3$.

\begin{table*}[t!]
   \vspace{10pt}
  \centering
  \resizebox{\textwidth}{!}
{
    \footnotesize
    \begin{tabular}{l||c|c||c|c|c|c|c|c|c}
  \toprule
    
 & \multirow{2}{*}{\begin{tabular}{@{}c@{}}Weighted \\ multi-scale\end{tabular}} 
 & \multirow{2}{*}{\begin{tabular}{@{}c@{}}Outlier \\ masking\end{tabular}} 
 & \multicolumn{4}{c|}{Error metric $\downarrow	$} & \multicolumn{3}{c}{Accuracy metric $\uparrow  $}\\
 \cline{4-7}
  \cline{8-10}
   &&   & Abs Rel & Sq Rel & RMSE  & RMSE log & $\delta < 1.25 $ & $\delta < 1.25^{2}$ & $\delta < 1.25^{3}$   \\
  \hline
  \hline
 Baseline & &  &  0.120  &   0.927  &   4.938  &   0.198  &   0.868  &   0.956  &   0.980  \\
\clinegrayone
 w/o  Weighted multi-scale & & \checkmark &   0.117  &   0.921  &   4.915  &   0.196  &   0.871  &   0.957  &   {\bf 0.981}  \\
\clinegrayone
 w/o Outlier masking & \checkmark &  &   0.115  &   0.910  &   4.865  &   0.193  &   0.876  &   0.958  &   0.980  \\
\clinegrayone
 \textbf{DiPE} & \checkmark & \checkmark & {\bf 0.112} &   {\bf 0.875} &   {\bf 4.795} &   {\bf 0.190} &   {\bf 0.880} &   {\bf 0.960} &   {\bf 0.981} \\ 
 
\bottomrule
\end{tabular}
}
    \vspace{-3pt}
      \caption{
      \textbf{Ablation Experiments.} There are 4 model variants with monocular training on the Eigen split \cite{eigen2014depth} of the KITTI dataset \cite{geiger2013vision}. 
The baseline model adopts existing principled masking, auto-masking and minimum reprojection techniques. Other 3 models include either one or both of our two contributions, the outlier making and weighted multi-scale methods. }
\label{tab:kitti_eigen_ablation}
\end{table*}

\subsubsection{Performance Comparison}
We quantitatively and qualitatively compare the results of our model and other state-of-the-art methods. 
The quantitative results are shown in Table~\ref{tab:kitti_eigen} and the results of other methods are taken from the corresponding papers. 
The comparison is mainly among the unsupervised monocular training methods, but some representative depth supervised, and unsupervised stereo training models are also included. 
DiPE archives state-of-the-art performance, as it markedly outperforms the current state of the art, Monodepth2 \cite{godard2019digging}. 
Also, DiPE has a comparable or even better performance to the models of the other two categories. 
Fig~\ref{fig:comparison} demonstrates the qualitative comparison among the predicted depth maps by DiPE and many state-of-the-art unsupervised monocular training methods. The predicted depth maps of other models are either shared by the authors or obtained by running the codes provided by the authors. DiPE handles the the scene dynamics and artifacts better than other methods. More results between DiPE and Monodepth2 \cite{godard2019digging} about oncoming vehicles can be seen from the attached video \url{https://youtu.be/UH8f-WkxVmU}.

\subsubsection{Ablation Study}

We also perform ablation experiments to examine the effectiveness of our contributions. 
As mentioned in Section \ref{sec:objective}, the baseline model uses the three existing masking techniques, i.e., the principled masking, auto-masking, and minimum reprojection. We experiment with four possible combinations of whether including our two contributions, the weighted multi-scale scheme, and the outlier masking technique. 
The results are shown in Table~\ref{tab:kitti_eigen_ablation}. 

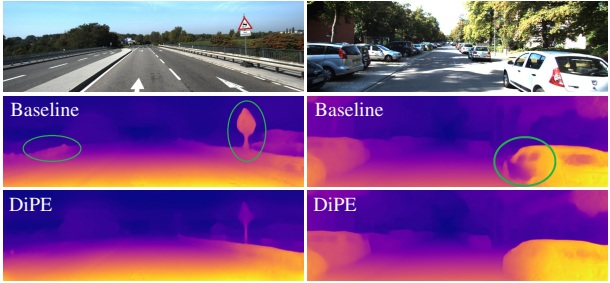
\begin{figure}
\centering
\hspace{-10pt}
\resizebox{\linewidth}{!}{
\input{figs/artifacts/artifacts.tex}}
\caption{{\bf Artifacts.} DiPE can solve the artifacts better and success in the two failure cases by Monodepth2 as is reported in the paper of Monodepth2 \cite{godard2019digging}. }
\vspace{-2pt}
\label{fig:artifacts}
\end{figure}

It can be observed that, our two contributions can obviously improve the performance individually, and the performance gain when they combine together is more than double of their separate performance gain, which indicates that the two techniques can collaborate well. Furthermore, the weighted multi-scale scheme also helps DiPE address the artifacts better than Monodepth2 \cite{godard2019digging}. DiPE can handle the two failure cases in Monodepth2, as illustrated in Fig.~\ref{fig:artifacts}.

\subsection{KITTI Odometry}
To prove the effectiveness of our DiPE model in the ego-motion estimation, we also experiment on the official odometry split of the KITTI dataset \cite{geiger2013vision}. We use three different input settings for the ego-motion network, with the number of frames as $2$, $3$, and $5$, respectively. For training the ego-motion network with input as $2$ or $3$ frames, we use the DiPE based on the baseline model. However, for the 5-frame-input network, we do not adopt the minimum reprojection technique, because it almost masks out all the pixels from the source views with indexes of $t-2$ and $t+2$ and the motion estimation for these two views is inferior.

\begin{table}[!ht]
\centering
\resizebox{1.0\columnwidth}{!}{
\begin{tabular}{l|c|c|c}
 \toprule
Method & \textbf{Sequence 09}     & \textbf{Sequence 10} & \textbf{\# frames} \\ \hline
\multicolumn{1}{l|}{ORB-SLAM \cite{mur2015orb}} & 0.014$\pm$0.008 & 0.012$\pm$0.011 & -  \\ \hline \hline
\multicolumn{1}{l|}{SfMLearner \cite{zhou2017unsupervised}} & 0.021$\pm$0.017 & 0.020$\pm$0.015 & 5\\ 
\multicolumn{1}{l|}{DF-Net \cite{zou2018df}} & 0.017$\pm$0.007 & 0.015$\pm$0.009 & 5 \\ 
\multicolumn{1}{l|}{GeoNet \cite{yin2018geonet}} & {0.012}$\pm$0.007 & {0.012$\pm$0.009} & 5 \\  
\multicolumn{1}{l|}{\textbf{DiPE} (Ours)} &\textbf{0.012$\pm$0.006} &\textbf{0.012$\pm$0.008}& 5 \\ 
\hline
\hline
\multicolumn{1}{l|}{DDVO \cite{wang2018learning}} & 0.045$\pm$0.108 & 0.033$\pm$0.074 & 3 \\ 
\multicolumn{1}{l|}{Vid2Depth \cite{mahjourian2018unsupervised}} & 0.013$\pm$0.010 & {0.012}$\pm$0.011 & 3 \\ 
\multicolumn{1}{l|}{EPC++ \cite{luo2019every}} & 0.013$\pm$0.007 & \textbf{0.012$\pm${0.008}} & 3\\ 
\multicolumn{1}{l|}{\textbf{DiPE} (Ours)} & \textbf{0.012$\pm$0.006} & \textbf{0.012$\pm$0.008} & 3 \\ 
\hline 
\hline
\multicolumn{1}{l|}{{Monodepth2 \cite{godard2019digging}}} & 0.017$\pm$0.008 & 0.015$\pm$0.010 & 2 \\ 
\multicolumn{1}{l|}{\textbf{DiPE} (Ours)} & \textbf{0.013$\pm$0.006} & \textbf{0.012$\pm$0.008}  & 2 \\ 
\bottomrule
\end{tabular}
}
\vspace{1pt}
\caption{
\textbf{Visual odometry results on the odometry split of the KITTI \cite{geiger2013vision} dataset.} Results show the average absolute trajectory error, and standard deviation, in meters.}
\label{tab:odom}
\end{table}

For evaluation, we adopt the commonly used metric proposed by Zhou \etal~\cite{zhou2017unsupervised}, i.e., the Absolute Trajectory Error (ATE)~\cite{mur2015orb} in $5$-frame snippets. The results are shown in Table~\ref{tab:odom} and the results of other models are taken from their corresponding papers. Among models with the three different input settings, DiPE achieves the best performance. Notably, in the setting of the pair-input ego-motion network, DiPE significantly outperforms Monodepth2 \cite{godard2019digging}. Besides, there is no significant performance difference among different motion network settings for DiPE, so DiPE is robust to different motion network input settings.

\section{CONCLUSION}
In this paper, we have demonstrated that carefully processing the photometric errors for unsupervised learning of depth and ego-motion from monocular videos can successfully solve the intrinsic difficulties, i.e., the occlusion and scene dynamics. 
We have introduced the outlier masking technique to exclude the irregular photometric errors that may mislead the network learning. This technique is useful in tackling occlusion and scene dynamics, especially for contra-directionally moving objects. Moreover, we have proposed an efficient and effective weighted multi-scale scheme to avoid the artifacts brought by multi-scale training. 
Unlike other methods that introducing extra modules, our approaches are simple, as they can be very easily incorporated in the unsupervised geometry learning framework. 
We have experimentally proven the effectiveness of our two contributions and built a new state-of-the-art model, DiPE, on both monocular depth and ego-motion estimation.

{\small
\bibliographystyle{IEEEtran}
\bibliography{ref}
}

\end{document}

%% file: figs/comparison/comparison.tex
\newcommand{\turnheightnew}{0.176\columnwidth}

\centering

\renewcommand{\arraystretch}{0.5}
\begin{tabular}{@{\hskip 1mm}c@{\hskip 1mm}c@{\hskip 1mm}c@{\hskip 1mm}c@{\hskip 1mm}c@{}}

{\rotatebox{90}{\hspace{5mm}\scriptsize Input}} &
\includegraphics[height=\turnheightnew]{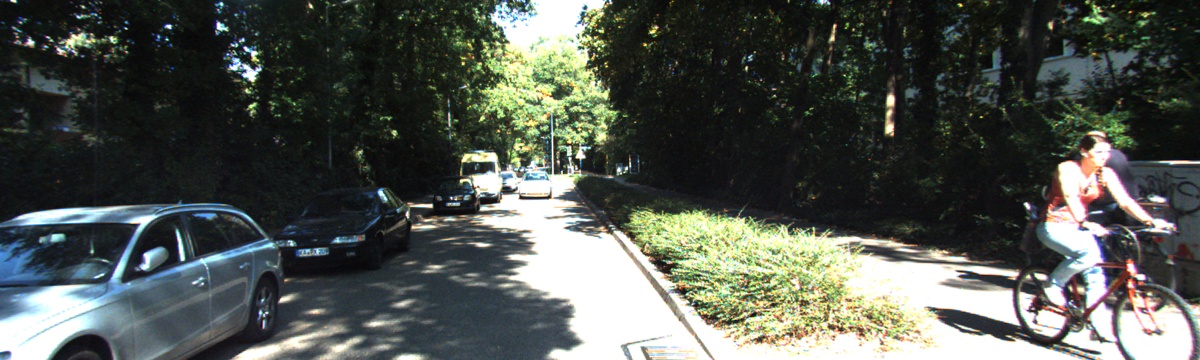} &
\includegraphics[height=\turnheightnew]{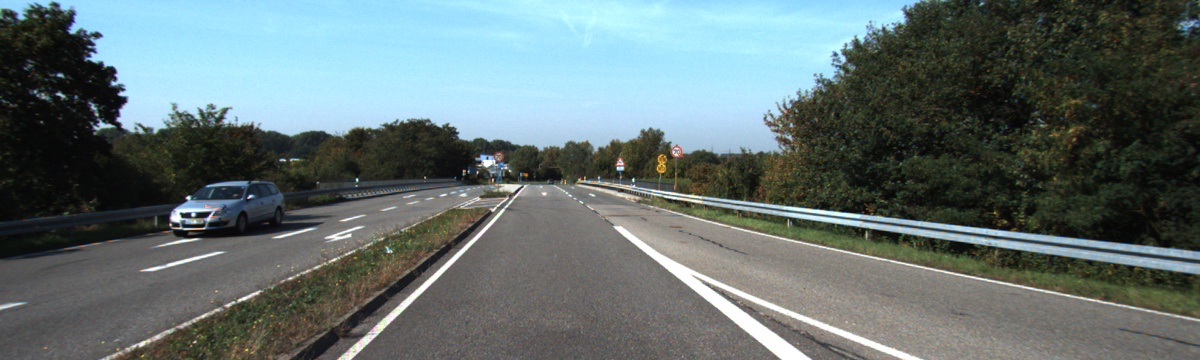} &
\includegraphics[height=\turnheightnew]{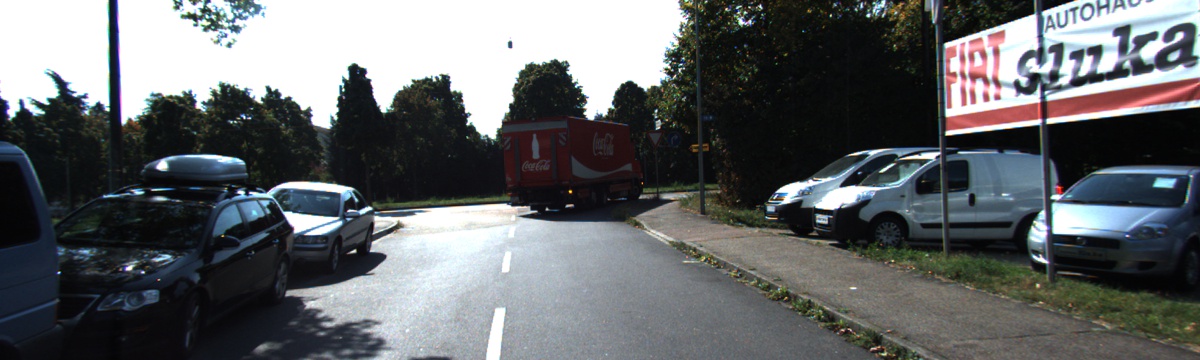} &
\includegraphics[height=\turnheightnew]{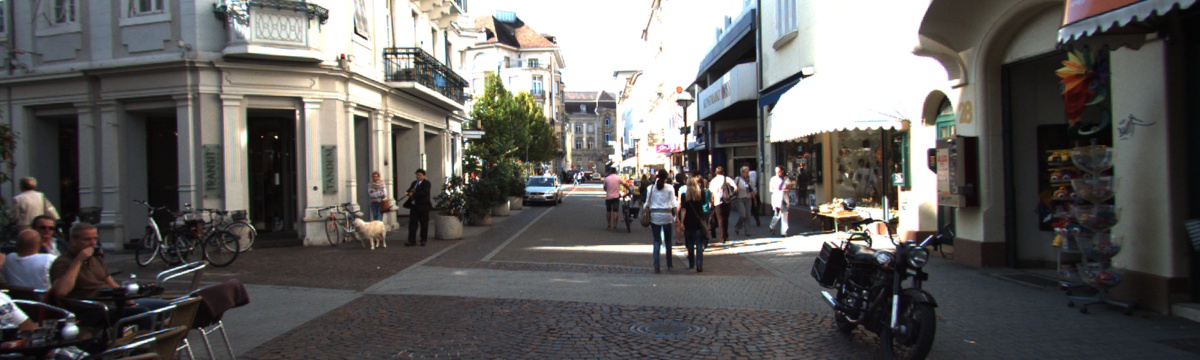}\\

{\rotatebox{90}{\hspace{0mm}\scriptsize
{Ground Truth}}} &
\includegraphics[height=\turnheightnew]{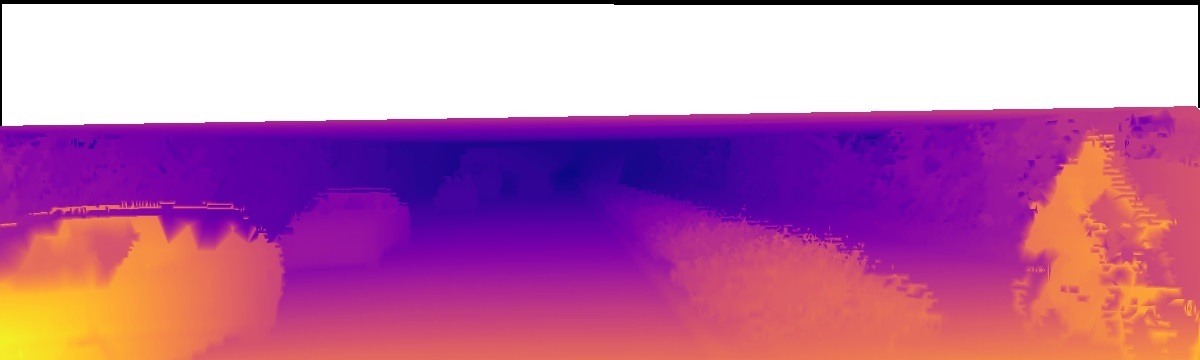} &
\includegraphics[height=\turnheightnew]{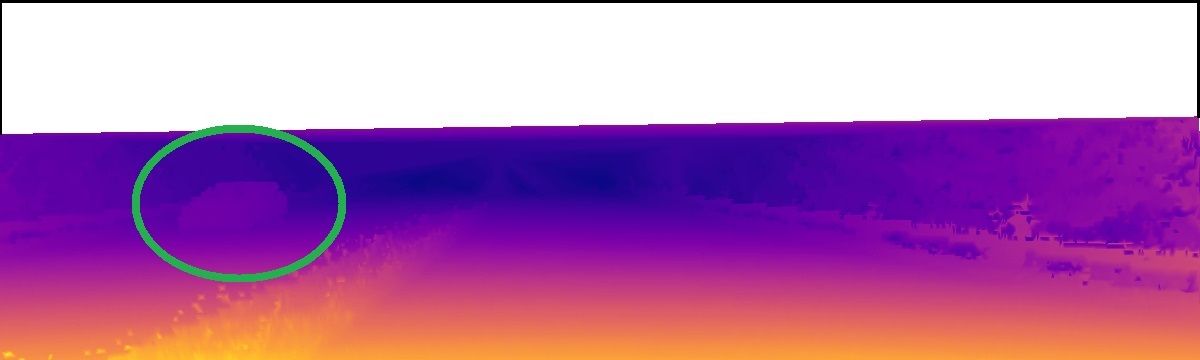} &
\includegraphics[height=\turnheightnew]{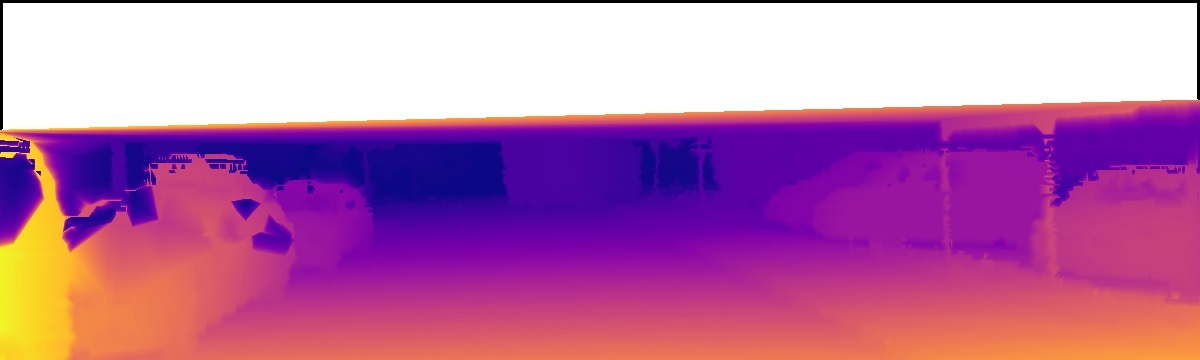} &
\includegraphics[height=\turnheightnew]{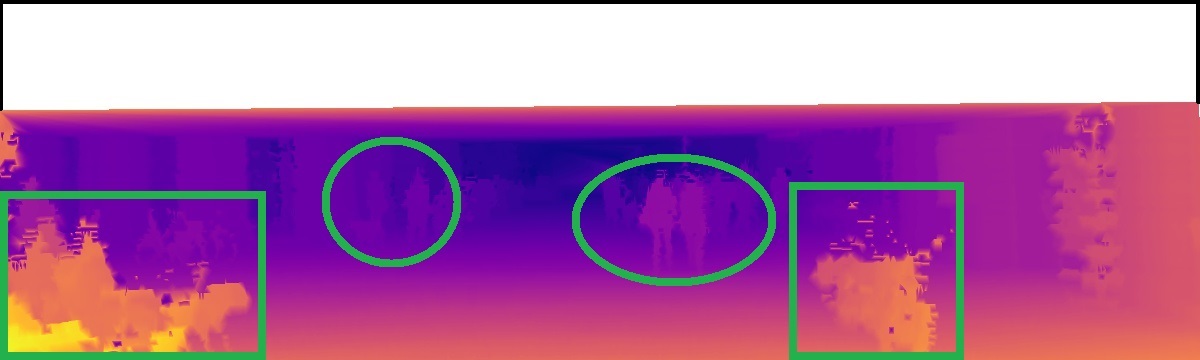}\\

{\rotatebox{90}{\hspace{0mm}\scriptsize
{SfMLearner\cite{zhou2017unsupervised}}}} &
\includegraphics[height=\turnheightnew]{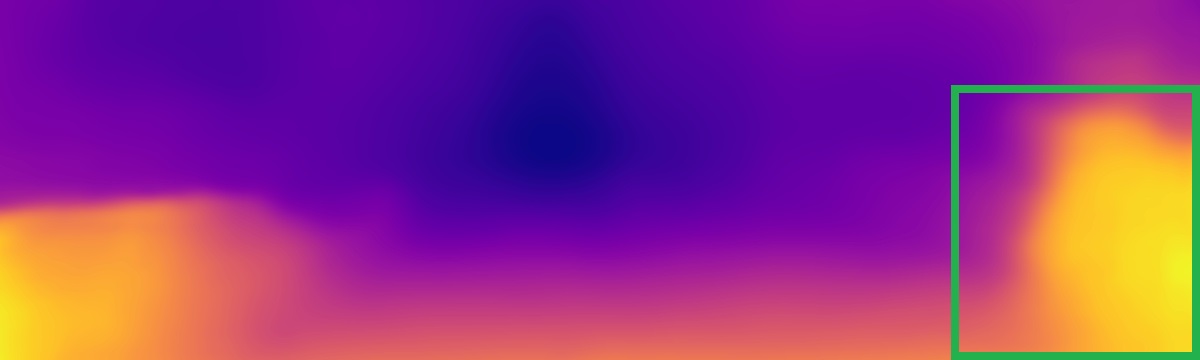} &
\includegraphics[height=\turnheightnew]{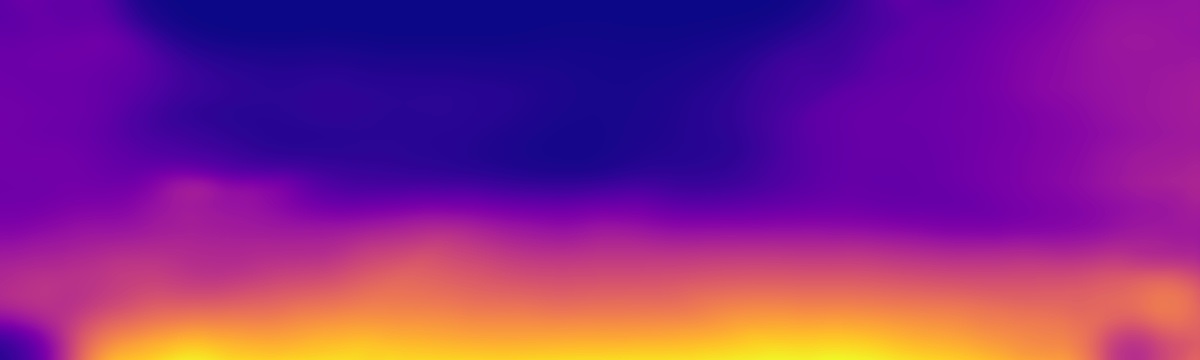} &
\includegraphics[height=\turnheightnew]{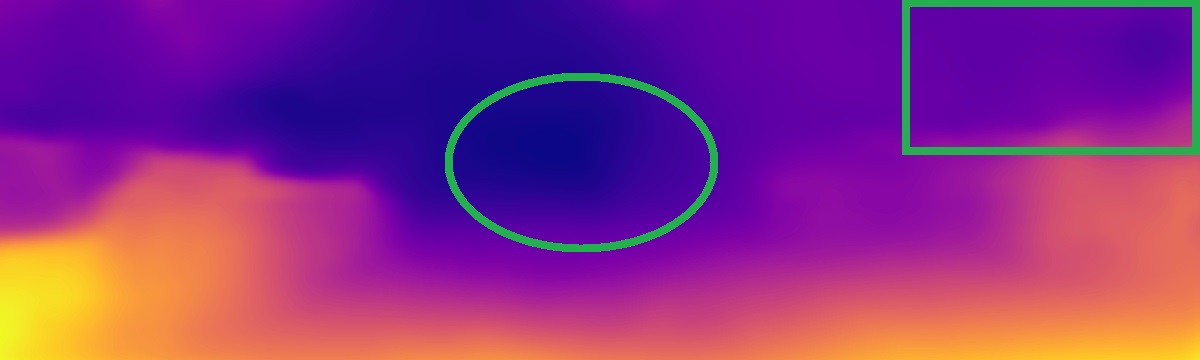} &
\includegraphics[height=\turnheightnew]{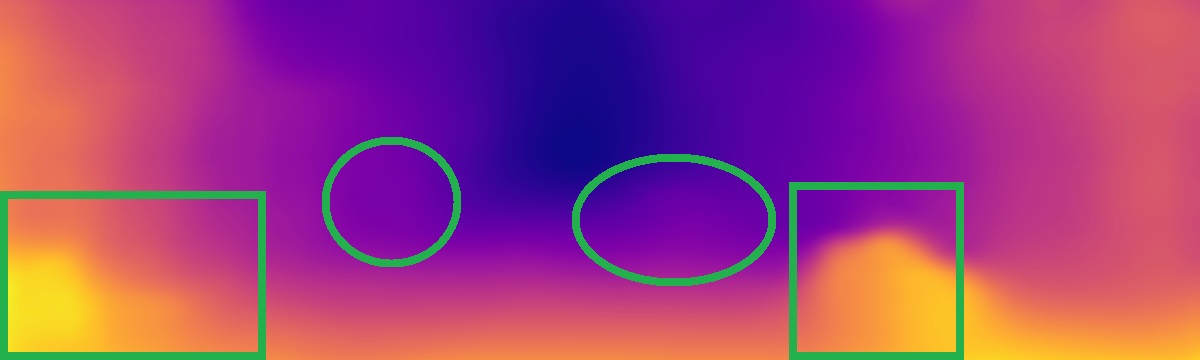}\\

{\rotatebox{90}{\hspace{0.5mm}\scriptsize
{ GeoNet\cite{yin2018geonet}}}} &
\includegraphics[height=\turnheightnew]{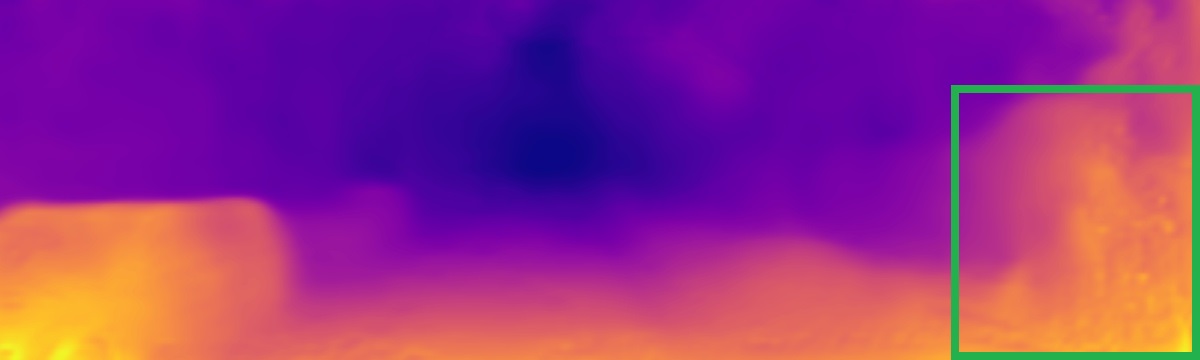} &
\includegraphics[height=\turnheightnew]{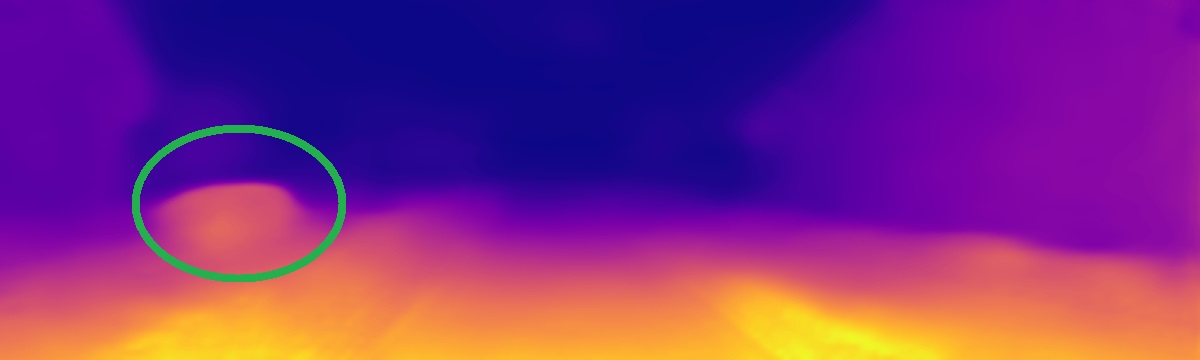} &
\includegraphics[height=\turnheightnew]{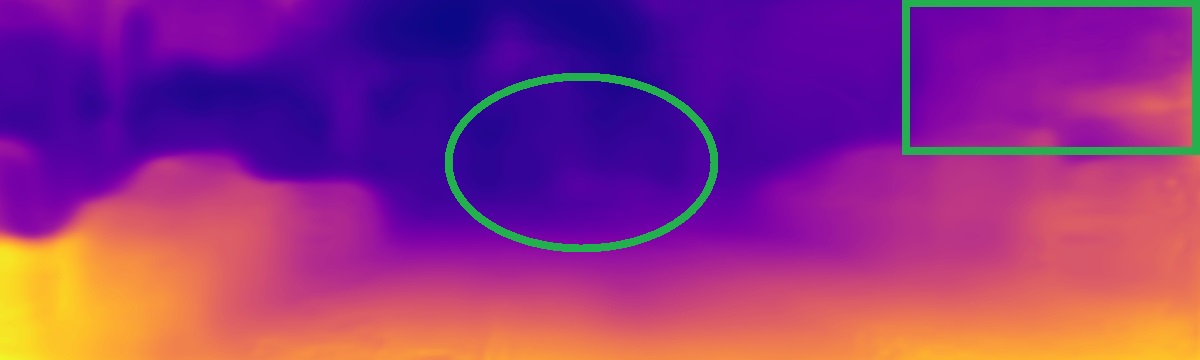} &
\includegraphics[height=\turnheightnew]{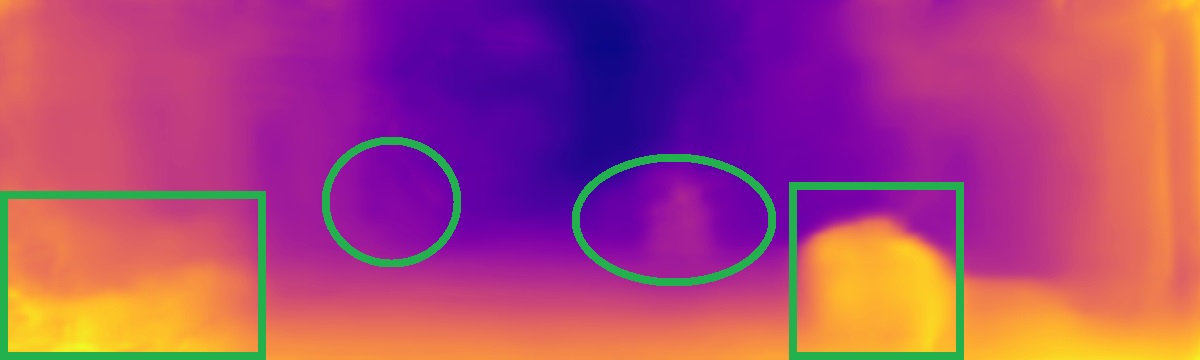}\\

{\rotatebox{90}{\hspace{1.5mm}\scriptsize
{ EPC++\cite{luo2019every}}}} &
\includegraphics[height=\turnheightnew]{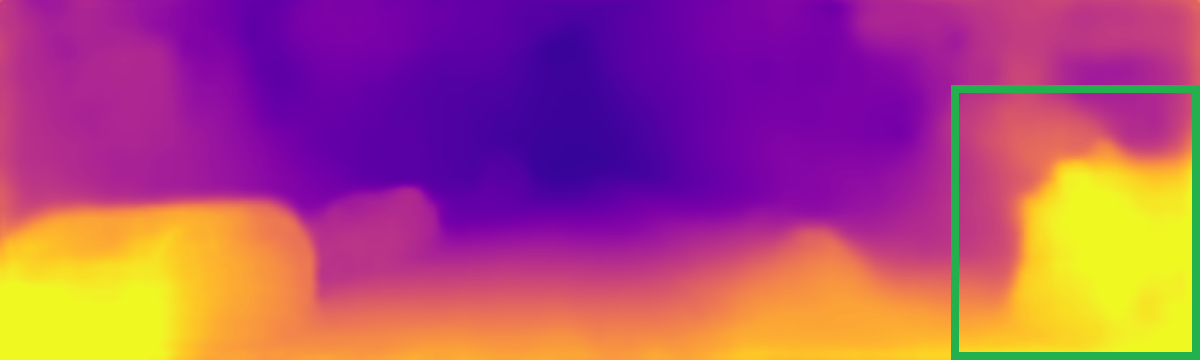} &
\includegraphics[height=\turnheightnew]{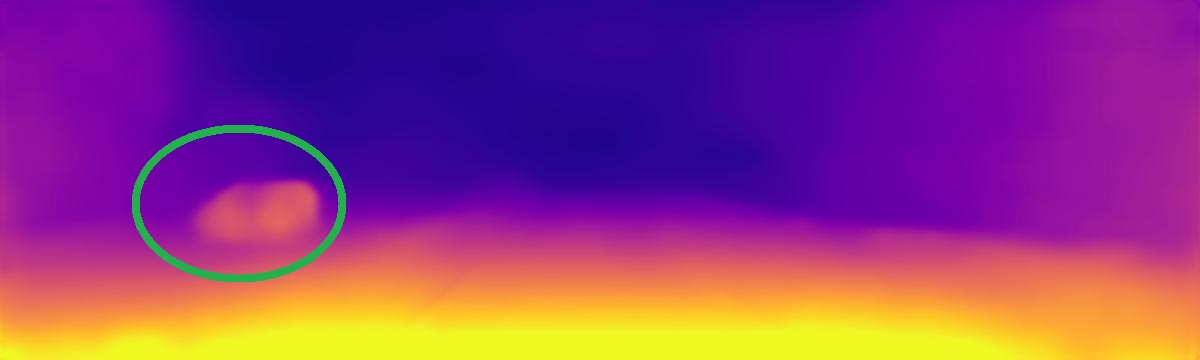} &
\includegraphics[height=\turnheightnew]{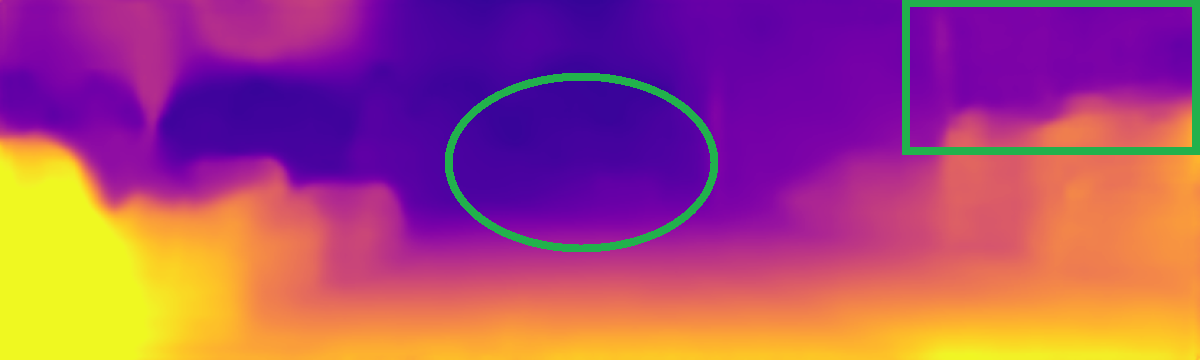} &
\includegraphics[height=\turnheightnew]{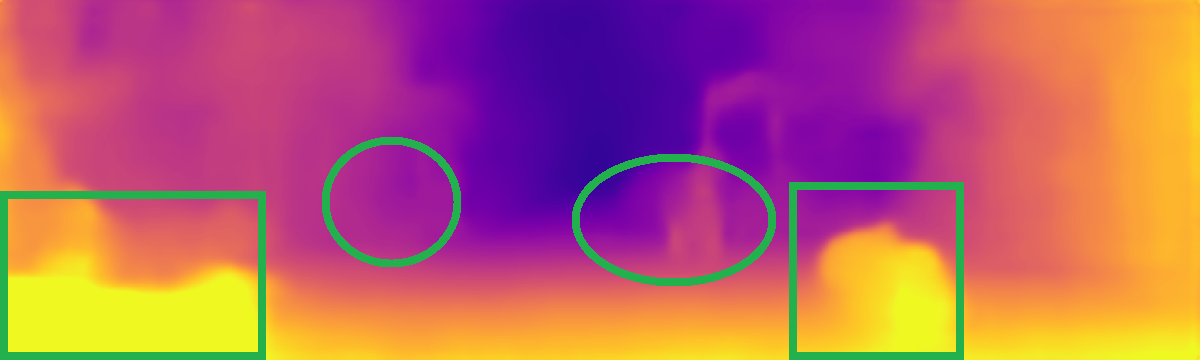}\\

{\rotatebox{90}{\hspace{0mm}\scriptsize
{Struct2depth\cite{casser2019struct2depth}}}} &
\includegraphics[height=\turnheightnew]{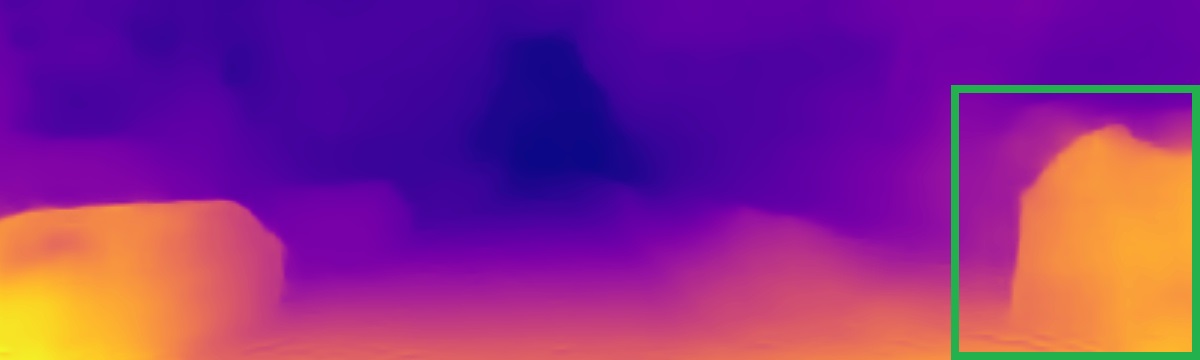} &
\includegraphics[height=\turnheightnew]{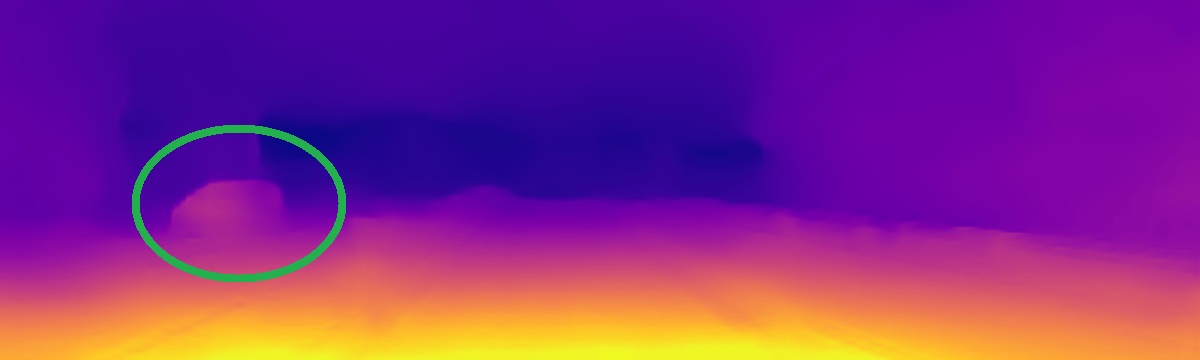} &
\includegraphics[height=\turnheightnew]{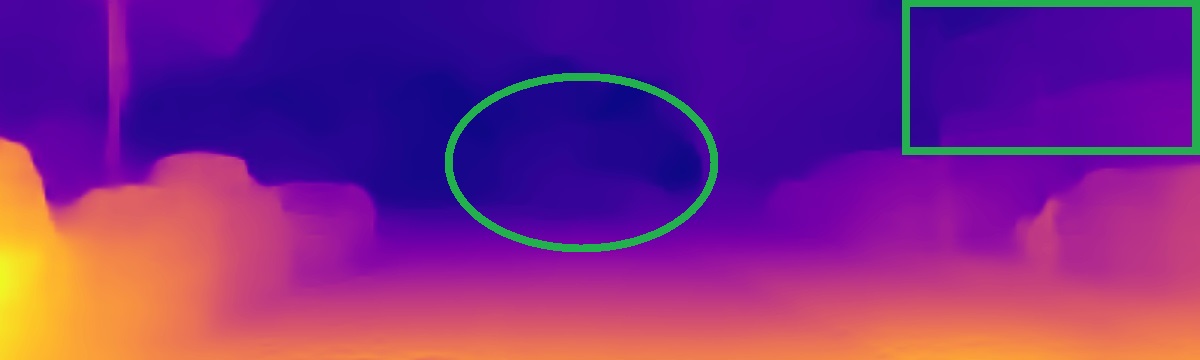} &
\includegraphics[height=\turnheightnew]{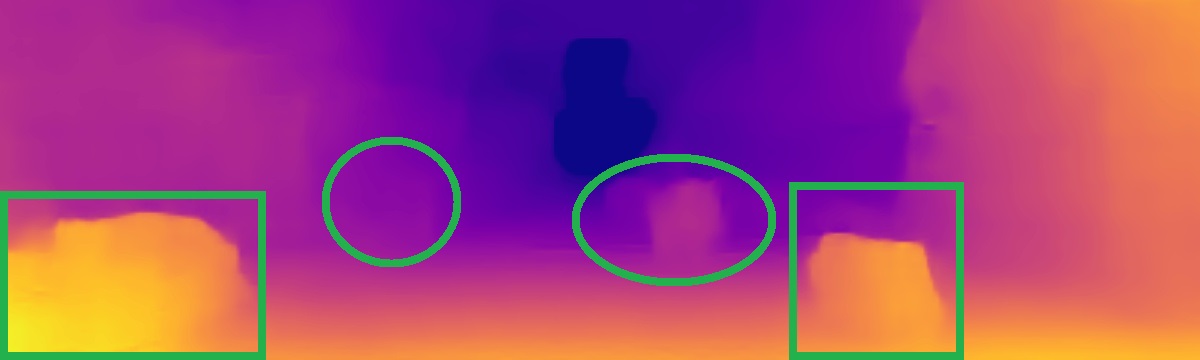}\\

{\rotatebox{90}{\hspace{0mm}\scriptsize
{Monodepth2\cite{godard2019digging}}}} &
\includegraphics[height=\turnheightnew]{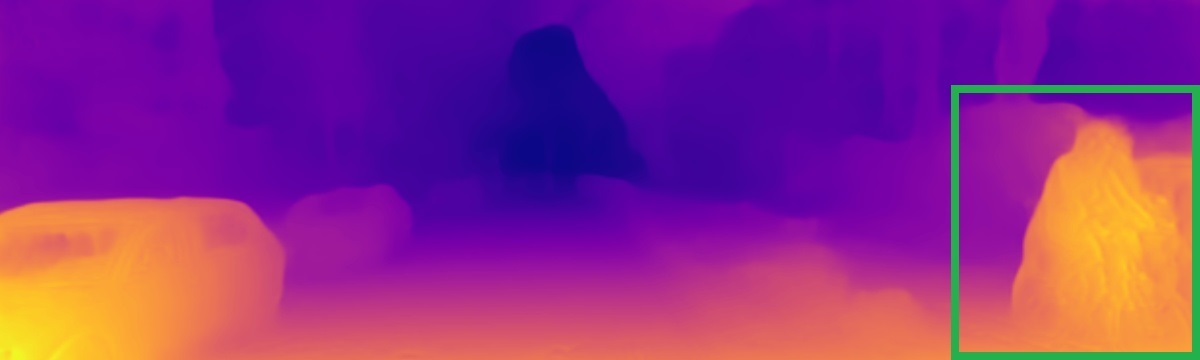} &
\includegraphics[height=\turnheightnew]{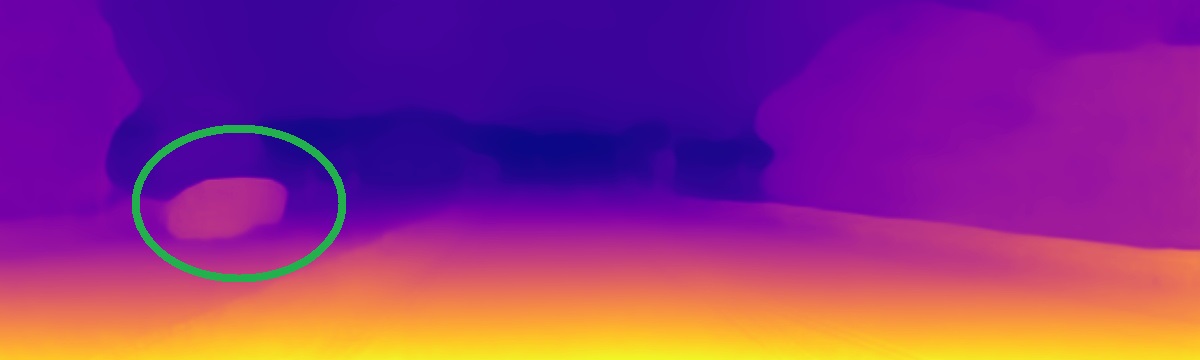} &
\includegraphics[height=\turnheightnew]{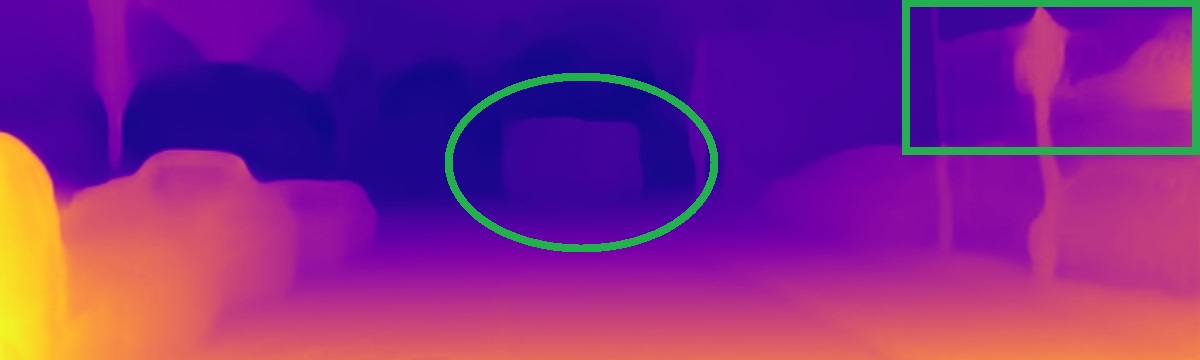} &
\includegraphics[height=\turnheightnew]{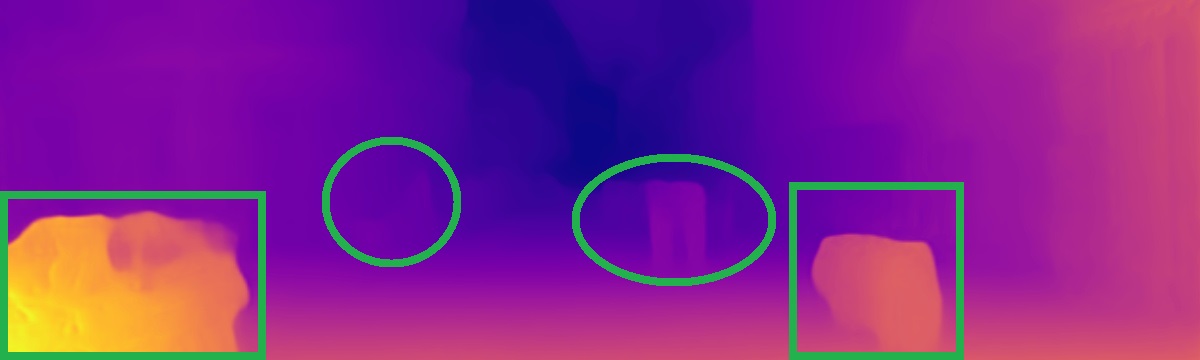}\\

{\rotatebox{90}{\hspace{1mm}\scriptsize
{DiPE(Ours)}}} &
\includegraphics[height=\turnheightnew]{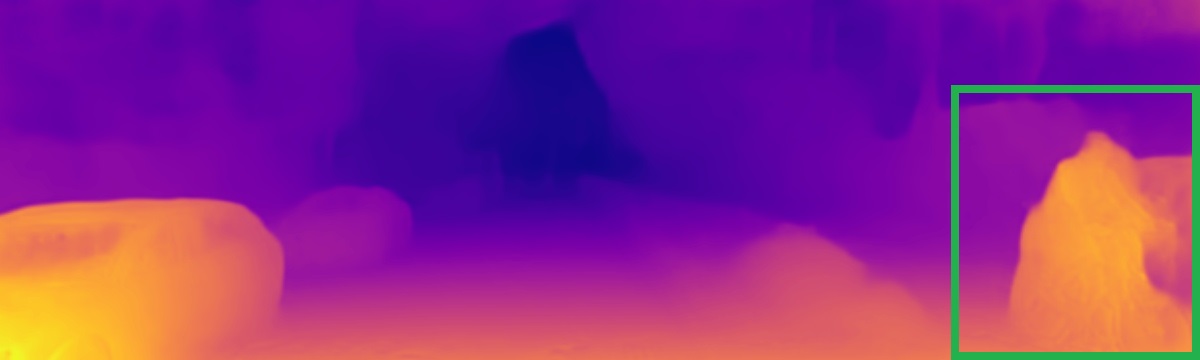} &
\includegraphics[height=\turnheightnew]{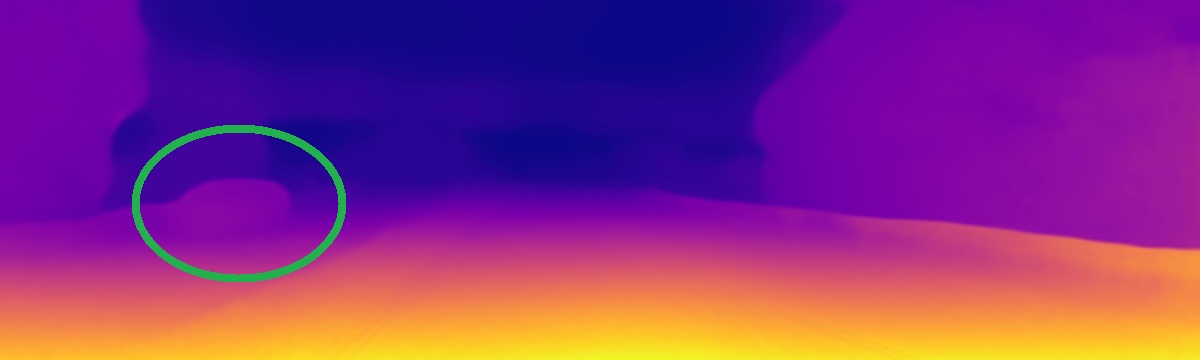} &
\includegraphics[height=\turnheightnew]{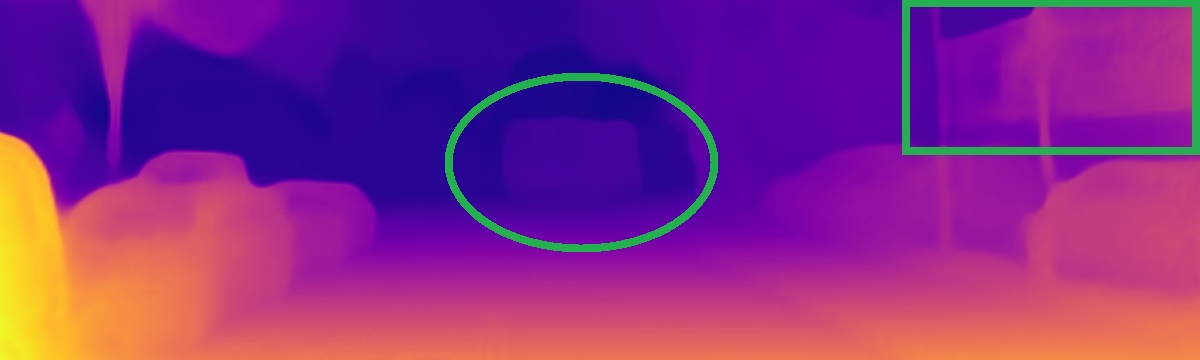} &
\includegraphics[height=\turnheightnew]{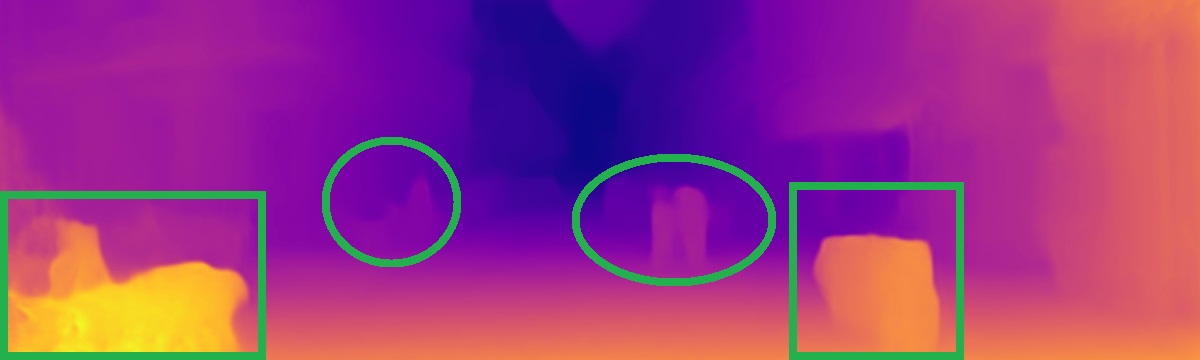}\\

\end{tabular}

%% file: figs/artifacts/artifacts.tex
\newcommand{\shiftleft}[2]{\makebox[-6pt][r]{\makebox[#1][l]{#2}}}
\newcommand{\imlabel}[2]{\includegraphics[width=0.49\columnwidth]{#1}%
\raisebox{28pt}{\shiftleft{82pt}{\makebox[-2pt][r]{\footnotesize #2}} }}

\centering
\renewcommand{\arraystretch}{0.5}
\begin{tabular}{@{\hskip -1mm}c@{\hskip 1.5mm}c}
\imlabel{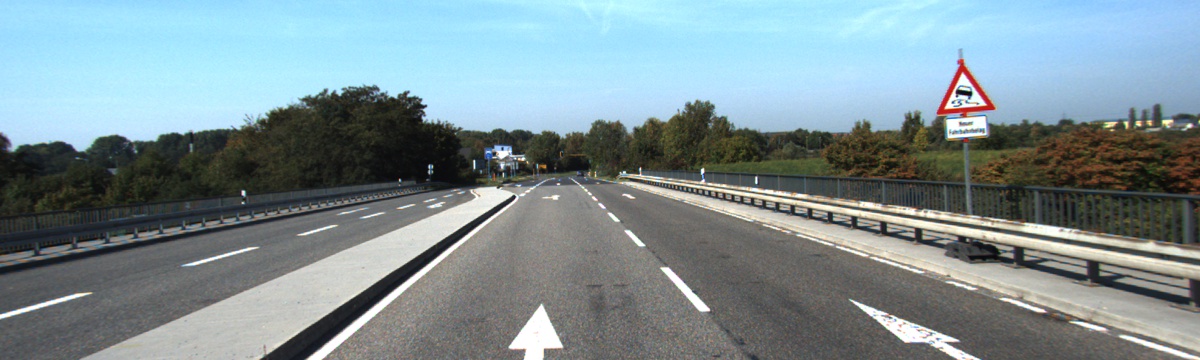}{} &
\imlabel{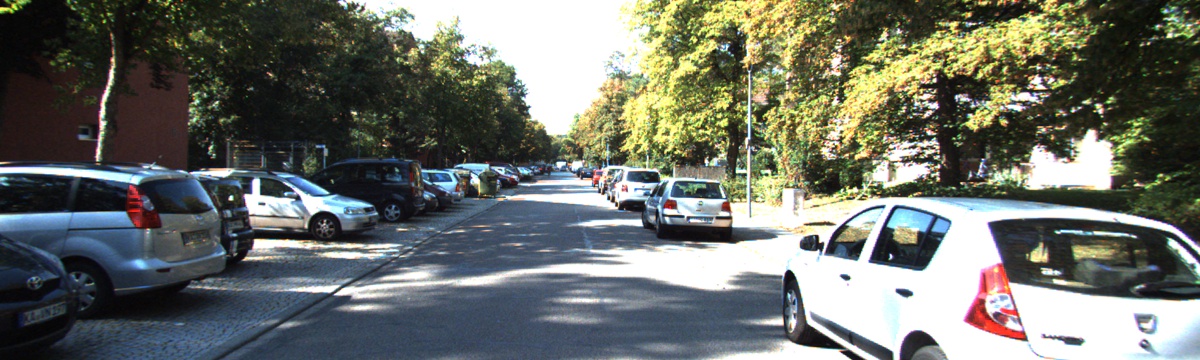}{} \\
\imlabel{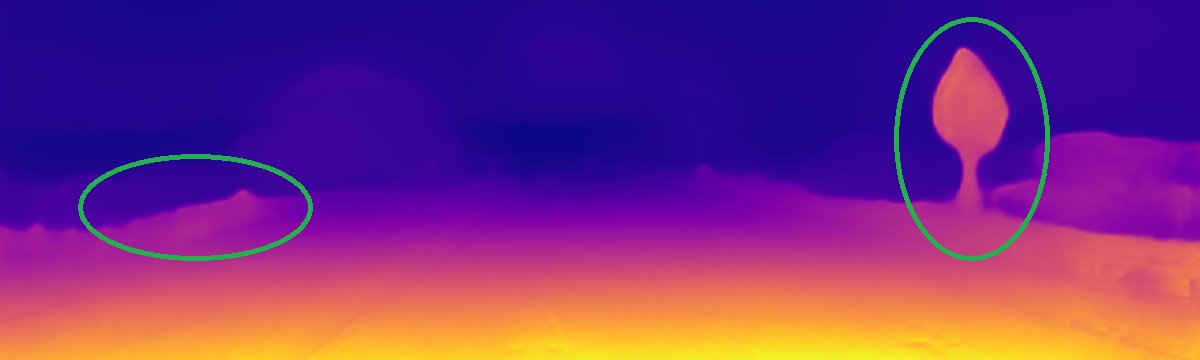}{\textcolor{white}{Baseline}} &
\imlabel{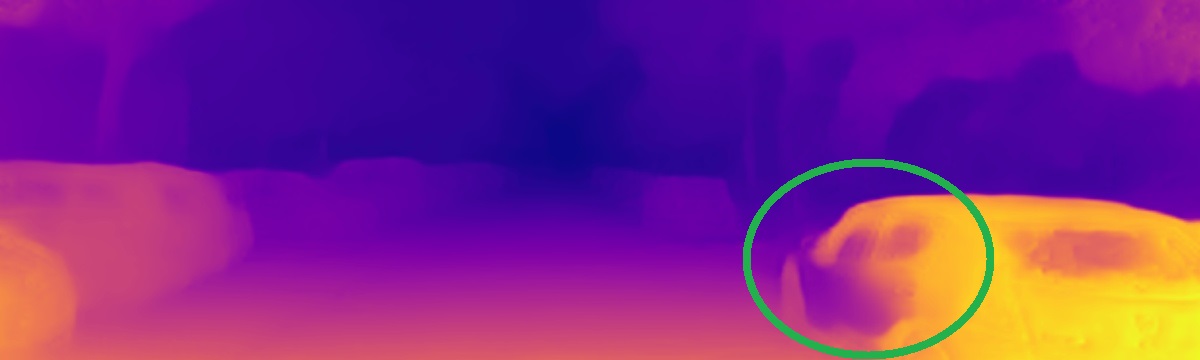}{\textcolor{white}{Baseline}} \\
\imlabel{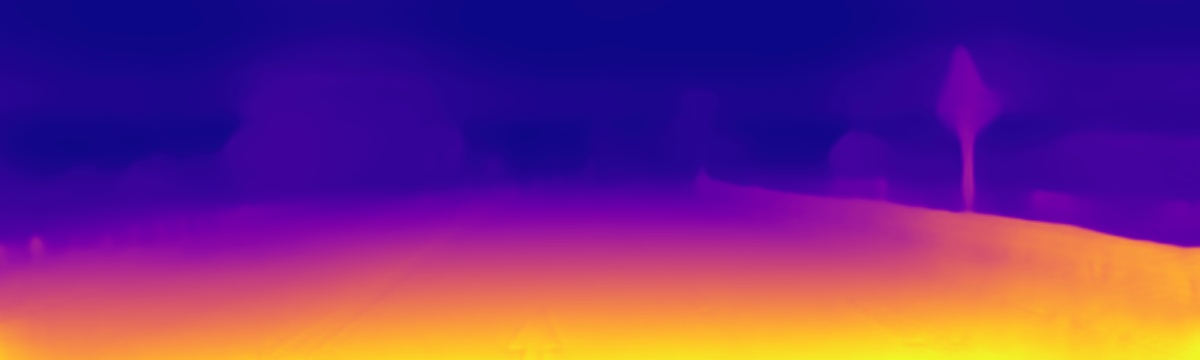}{\textcolor{white}{DiPE \quad }} &  
\imlabel{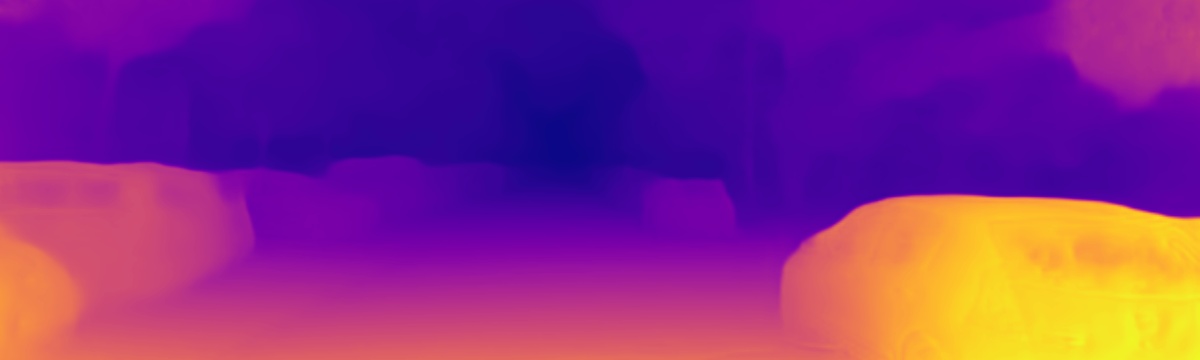}{\textcolor{white}{DiPE \quad }} \\

\end{tabular}

%% file: root.bbl
\begin{thebibliography}{10}
\providecommand{\url}[1]{#1}
\csname url@samestyle\endcsname
\providecommand{\newblock}{\relax}
\providecommand{\bibinfo}[2]{#2}
\providecommand{\BIBentrySTDinterwordspacing}{\spaceskip=0pt\relax}
\providecommand{\BIBentryALTinterwordstretchfactor}{4}
\providecommand{\BIBentryALTinterwordspacing}{\spaceskip=\fontdimen2\font plus
\BIBentryALTinterwordstretchfactor\fontdimen3\font minus
  \fontdimen4\font\relax}
\providecommand{\BIBforeignlanguage}[2]{{%
\expandafter\ifx\csname l@#1\endcsname\relax
\typeout{** WARNING: IEEEtran.bst: No hyphenation pattern has been}%
\typeout{** loaded for the language `#1'. Using the pattern for}%
\typeout{** the default language instead.}%
\else
\language=\csname l@#1\endcsname
\fi
#2}}
\providecommand{\BIBdecl}{\relax}
\BIBdecl

\bibitem{zhou2017unsupervised}
T.~Zhou, M.~Brown, N.~Snavely, and D.~G. Lowe, ``Unsupervised learning of depth
  and ego-motion from video,'' in \emph{Proceedings of the IEEE Conference on
  Computer Vision and Pattern Recognition}, 2017, pp. 1851--1858.

\bibitem{luo2019every}
C.~Luo, Z.~Yang, P.~Wang, Y.~Wang, W.~Xu, R.~Nevatia, and A.~Yuille, ``Every
  pixel counts++: Joint learning of geometry and motion with 3d holistic
  understanding.'' \emph{IEEE transactions on pattern analysis and machine
  intelligence}, 2019.

\bibitem{casser2019struct2depth}
V.~Casser, S.~Pirk, R.~Mahjourian, and A.~Angelova, ``Depth prediction without
  the sensors: Leveraging structure for unsupervised learning from monocular
  videos,'' in \emph{Thirty-Third AAAI Conference on Artificial Intelligence
  (AAAI-19)}, 2019.

\bibitem{godard2019digging}
C.~Godard, O.~Mac~Aodha, M.~Firman, and G.~J. Brostow, ``Digging into
  self-supervised monocular depth estimation,'' in \emph{Proceedings of the
  IEEE International Conference on Computer Vision}, 2019, pp. 3828--3838.

\bibitem{eigen2014depth}
D.~Eigen, C.~Puhrsch, and R.~Fergus, ``Depth map prediction from a single image
  using a multi-scale deep network,'' in \emph{Advances in neural information
  processing systems}, 2014, pp. 2366--2374.

\bibitem{laina2016deeper}
I.~Laina, C.~Rupprecht, V.~Belagiannis, F.~Tombari, and N.~Navab, ``Deeper
  depth prediction with fully convolutional residual networks,'' in \emph{3D
  Vision (3DV), 2016 Fourth International Conference on}.\hskip 1em plus 0.5em
  minus 0.4em\relax IEEE, 2016, pp. 239--248.

\bibitem{zeng2017geocuedepth}
Y.~Zeng, Y.~Hu, S.~Liu, Q.~Tang, J.~Ye, and X.~Li, ``Geocuedepth: exploiting
  geometric structure cues to estimate depth from a single image,'' in
  \emph{2017 IEEE/RSJ International Conference on Intelligent Robots and
  Systems (IROS)}.\hskip 1em plus 0.5em minus 0.4em\relax IEEE, 2017, pp.
  17--22.

\bibitem{jiang2019high}
H.~Jiang and R.~Huang, ``High quality monocular depth estimation via a
  multi-scale network and a detail-preserving objective,'' in \emph{2019 IEEE
  International Conference on Image Processing (ICIP)}.\hskip 1em plus 0.5em
  minus 0.4em\relax IEEE, 2019, pp. 1920--1924.

\bibitem{wang2015towards}
P.~Wang, X.~Shen, Z.~Lin, S.~Cohen, B.~Price, and A.~L. Yuille, ``Towards
  unified depth and semantic prediction from a single image,'' in
  \emph{Proceedings of the IEEE Conference on Computer Vision and Pattern
  Recognition}, 2015, pp. 2800--2809.

\bibitem{liu2015learning}
F.~Liu, C.~Shen, G.~Lin, and I.~Reid, ``Learning depth from single monocular
  images using deep convolutional neural fields,'' \emph{IEEE transactions on
  pattern analysis and machine intelligence}, vol.~38, no.~10, pp. 2024--2039,
  2015.

\bibitem{xu2017multi}
D.~Xu, E.~Ricci, W.~Ouyang, X.~Wang, and N.~Sebe, ``Multi-scale continuous crfs
  as sequential deep networks for monocular depth estimation,'' in
  \emph{Proceedings of CVPR}, 2017.

\bibitem{cao2017estimating}
Y.~Cao, Z.~Wu, and C.~Shen, ``Estimating depth from monocular images as
  classification using deep fully convolutional residual networks,'' \emph{IEEE
  Transactions on Circuits and Systems for Video Technology}, 2017.

\bibitem{li2018monocular}
B.~Li, Y.~Dai, and M.~He, ``Monocular depth estimation with hierarchical fusion
  of dilated cnns and soft-weighted-sum inference,'' \emph{Pattern
  Recognition}, vol.~83, pp. 328--339, 2018.

\bibitem{fu2018deep}
H.~Fu, M.~Gong, C.~Wang, K.~Batmanghelich, and D.~Tao, ``Deep ordinal
  regression network for monocular depth estimation,'' in \emph{IEEE Conference
  on Computer Vision and Pattern Recognition (CVPR)}, 2018.

\bibitem{silberman2012indoor}
N.~Silberman, D.~Hoiem, P.~Kohli, and R.~Fergus, ``Indoor segmentation and
  support inference from rgbd images,'' in \emph{European Conference on
  Computer Vision}.\hskip 1em plus 0.5em minus 0.4em\relax Springer, 2012, pp.
  746--760.

\bibitem{geiger2013vision}
A.~Geiger, P.~Lenz, C.~Stiller, and R.~Urtasun, ``Vision meets robotics: The
  kitti dataset,'' \emph{The International Journal of Robotics Research},
  vol.~32, no.~11, pp. 1231--1237, 2013.

\bibitem{garg2016unsupervised}
R.~Garg, V.~K. BG, G.~Carneiro, and I.~Reid, ``Unsupervised cnn for single view
  depth estimation: Geometry to the rescue,'' in \emph{European Conference on
  Computer Vision}.\hskip 1em plus 0.5em minus 0.4em\relax Springer, 2016, pp.
  740--756.

\bibitem{godard2017unsupervised}
C.~Godard, O.~Mac~Aodha, and G.~J. Brostow, ``Unsupervised monocular depth
  estimation with left-right consistency,'' in \emph{Proceedings of the IEEE
  Conference on Computer Vision and Pattern Recognition}, 2017, pp. 270--279.

\bibitem{zhan2018unsupervised}
H.~Zhan, R.~Garg, C.~Saroj~Weerasekera, K.~Li, H.~Agarwal, and I.~Reid,
  ``Unsupervised learning of monocular depth estimation and visual odometry
  with deep feature reconstruction,'' in \emph{Proceedings of the IEEE
  Conference on Computer Vision and Pattern Recognition}, 2018, pp. 340--349.

\bibitem{li2018undeepvo}
R.~Li, S.~Wang, Z.~Long, and D.~Gu, ``Undeepvo: Monocular visual odometry
  through unsupervised deep learning,'' in \emph{2018 IEEE International
  Conference on Robotics and Automation (ICRA)}.\hskip 1em plus 0.5em minus
  0.4em\relax IEEE, 2018, pp. 7286--7291.

\bibitem{klodt2018supervising}
M.~Klodt and A.~Vedaldi, ``Supervising the new with the old: learning sfm from
  sfm,'' in \emph{Proceedings of the European Conference on Computer Vision
  (ECCV)}, 2018, pp. 698--713.

\bibitem{wang2019unsupervised}
G.~Wang, H.~Wang, Y.~Liu, and W.~Chen, ``Unsupervised learning of monocular
  depth and ego-motion using multiple masks,'' in \emph{2019 International
  Conference on Robotics and Automation (ICRA)}.\hskip 1em plus 0.5em minus
  0.4em\relax IEEE, 2019, pp. 4724--4730.

\bibitem{yin2018geonet}
Z.~Yin and J.~Shi, ``Geonet: Unsupervised learning of dense depth, optical flow
  and camera pose,'' in \emph{The IEEE Conference on Computer Vision and
  Pattern Recognition}, 2018, pp. 1983--1992.

\bibitem{zou2018df}
Y.~Zou, Z.~Luo, and J.-B. Huang, ``Df-net: Unsupervised joint learning of depth
  and flow using cross-task consistency,'' in \emph{Proceedings of the European
  Conference on Computer Vision (ECCV)}, 2018, pp. 36--53.

\bibitem{vijayanarasimhan2017sfm}
S.~Vijayanarasimhan, S.~Ricco, C.~Schmid, R.~Sukthankar, and K.~Fragkiadaki,
  ``Sfm-net: Learning of structure and motion from video,'' \emph{arXiv
  preprint arXiv:1704.07804}, 2017.

\bibitem{Gordon_2019_ICCV}
A.~Gordon, H.~Li, R.~Jonschkowski, and A.~Angelova, ``Depth from videos in the
  wild: Unsupervised monocular depth learning from unknown cameras,'' in
  \emph{The IEEE International Conference on Computer Vision (ICCV)}, October
  2019.

\bibitem{jaderberg2015spatial}
M.~Jaderberg, K.~Simonyan, A.~Zisserman \emph{et~al.}, ``Spatial transformer
  networks,'' in \emph{Advances in neural information processing systems},
  2015, pp. 2017--2025.

\bibitem{wang2018learning}
C.~Wang, J.~Miguel~Buenaposada, R.~Zhu, and S.~Lucey, ``Learning depth from
  monocular videos using direct methods,'' in \emph{Proceedings of the IEEE
  Conference on Computer Vision and Pattern Recognition}, 2018, pp. 2022--2030.

\bibitem{mahjourian2018unsupervised}
R.~Mahjourian, M.~Wicke, and A.~Angelova, ``Unsupervised learning of depth and
  ego-motion from monocular video using 3d geometric constraints,'' in
  \emph{Proceedings of the IEEE Conference on Computer Vision and Pattern
  Recognition}, 2018, pp. 5667--5675.

\bibitem{kuznietsov2017semi}
Y.~Kuznietsov, J.~St{\"u}ckler, and B.~Leibe, ``Semi-supervised deep learning
  for monocular depth map prediction,'' in \emph{The IEEE Conference on
  Computer Vision and Pattern Recognition}, 2017, pp. 6647--6655.

\bibitem{pillai2019superdepth}
S.~Pillai, R.~Ambru{\c{s}}, and A.~Gaidon, ``Superdepth: Self-supervised,
  super-resolved monocular depth estimation,'' in \emph{2019 International
  Conference on Robotics and Automation (ICRA)}.\hskip 1em plus 0.5em minus
  0.4em\relax IEEE, 2019, pp. 9250--9256.

\bibitem{he2016deep}
K.~He, X.~Zhang, S.~Ren, and J.~Sun, ``Deep residual learning for image
  recognition,'' in \emph{Proceedings of the IEEE conference on computer vision
  and pattern recognition}, 2016, pp. 770--778.

\bibitem{deng2009imagenet}
J.~Deng, W.~Dong, R.~Socher, L.-J. Li, K.~Li, and L.~Fei-Fei, ``Imagenet: A
  large-scale hierarchical image database,'' in \emph{2009 IEEE conference on
  computer vision and pattern recognition}.\hskip 1em plus 0.5em minus
  0.4em\relax Ieee, 2009, pp. 248--255.

\bibitem{kingma2014adam}
D.~P. Kingma and J.~Ba, ``Adam: A method for stochastic optimization,''
  \emph{arXiv preprint arXiv:1412.6980}, 2014.

\bibitem{mur2015orb}
R.~Mur-Artal, J.~M.~M. Montiel, and J.~D. Tardos, ``Orb-slam: a versatile and
  accurate monocular slam system,'' \emph{IEEE transactions on robotics},
  vol.~31, no.~5, pp. 1147--1163, 2015.

\end{thebibliography}
